\documentclass[pmlr]{jmlr}

\usepackage{amsmath,amsfonts,bm}

\def\eqref#1{equation~\ref{#1}}

\def\1{\bm{1}}

\DeclareMathAlphabet{\mathsfit}{\encodingdefault}{\sfdefault}{m}{sl}
\SetMathAlphabet{\mathsfit}{bold}{\encodingdefault}{\sfdefault}{bx}{n}

\usepackage[T1]{fontenc}
\usepackage{booktabs}
\usepackage{microtype}
\usepackage{xurl}
\numberwithin{theorem}{section}
\theorembodyfont{\normalfont\upshape}

\newtheorem{definition}[theorem]{Definition}
\newtheorem{assumption}[theorem]{Assumption}
\theoremheaderfont{\normalfont\itshape}

\newtheorem{remark}[theorem]{Remark}
\theorembodyfont{\normalfont\itshape}
\theoremheaderfont{\normalfont\bfseries}
\renewcommand{\eqref}[1]{\textup{(\ref{#1})}}

\jmlrvolume{}
\jmlryear{}
\jmlrworkshop{}

\title[User-Centered Agentic Recommendation]{Toward User-Centered Agentic Recommendation: When Does Platform Competition Benefit Users?}
\hypersetup{
  pdftitle={Toward User-Centered Agentic Recommendation: When Does Platform Competition Benefit Users?},
  pdfauthor={Kehan Zheng, Deyao Hong, Qian Li, Jun Zhang, Jie Jiang, Hongning Wang},
  pdfkeywords={user-centered agentic recommendation, recommender system, welfare analysis, game theory}
}

\author[Zheng et al.]{%
  \Name{Kehan Zheng\nametag{$^{1}$\thanks{Equal contribution.}}},
  \Name{Deyao Hong\nametag{$^{1}$\footnotemark[1]}},
  \Name{Qian Li\nametag{$^{2}$}}\\
  \Name{Jun Zhang\nametag{$^{2}$}},
  \Name{Jie Jiang\nametag{$^{2}$}},
  \Name{Hongning Wang\nametag{$^{1}$\thanks{Corresponding author.}}}\\
  \addr $^{1}$Tsinghua University \qquad $^{2}$Tencent Inc.\\
  \upshape
  \Email{zhengkh24@mails.tsinghua.edu.cn},
  \Email{hongdy22@mails.tsinghua.edu.cn},
  \Email{hw-ai@tsinghua.edu.cn}
}

\begin{document}

\maketitle

\begin{abstract}
Conventional recommender systems operate within platform-specific catalogs, limiting access to potentially valuable resources across platforms.
We propose user-centered agentic recommendation, a paradigm in which user agents compare recommendations from competing platform agents, and formalize their interaction as a repeated game.
Platforms strategically modify descriptions without changing intrinsic item value, while each user agent assesses platform credibility using feedback from consumed items only.
Our analysis shows that expanded resource access need not improve user welfare: strategic misrepresentation and limited verification can offset the benefits of platform competition.
We quantify when platform credibility assessment allows users to benefit from broader access despite strategic exaggeration. Our analysis identifies resource and reporting conditions that guarantee higher average welfare than the original platform with an interaction budget linear in platform count.
Experiments on synthetic and Amazon-grounded markets show that credibility assessment can improve finite-horizon utility, while its benefits depend on market size and mechanism parameters.
\end{abstract}


\section{Introduction}
\label{sec:introduction}

Most recommender systems organize choice within a platform: they learn from user interactions, rank items in their own catalogs, and collect feedback on the resulting recommendations. Finding alternatives across platforms remains the user's responsibility. Advances in large language models (LLMs) and agents suggest a different organization. Recommendation agents combine planning and tool use~\citep{wang2024recmind}, and can interpret personal information need and mediate interactions with recommenders~\citep{xu2025iagent}, motivating recent proposals for user-centered services and personalization beyond platform boundaries~\citep{zhang2026usercentric,lin2026usergoverned}.

In this work, we propose \emph{user-centered agentic recommendation}: a personal user agent solicits recommendations from multiple platform agents to find the most relevant resource on the user's behalf. Each platform retains its private catalog and responds independently. The user agent brings these responses together at the point of choice, without requiring competitors to share catalogs or a central service to control their resources. Recommendation thus moves from ranking within a platform already chosen by the user to comparing what different platforms offer for the same request.

This expanded access creates a new difficulty. Platforms competing for selection can improve how a resource is described without changing what the user actually receives. 
Platform credibility assessment offers a way to discount such mismatch, but each interaction can only verify the consumed resource, echoing the information problem of experience goods~\citep{nelson1970information}. Adding platforms expands the set of assessments needed without increasing the feedback available to each user; strategically changing the descriptions can also make past assessments less useful. This raises a central question: \emph{When does user-centered agentic recommendation benefit users compared with staying on a single platform?}

We model repeated interactions between users with different interests and platforms offering different resources. Platforms compete for exposure by changing the descriptions of their recommended resources, not the resources themselves. Correspondingly, each user agent learns from the target user's consumption and uses this experience to assess the credibility of the platforms. We compare the utility users receive when staying on their original platform, and also against the situation when accurate item description is available. 

Without the platform's credibility assessment on the user agent side, there is a weak Nash equilibrium in which every platform claims the highest possible relevance, so that users cannot distinguish better resources through the descriptions. Credibility assessment limits this loss. Our welfare bound separates the cost of countering platforms' strategic overstatement. Most importantly, this allows user side tolerance of adaptive descriptions, provided that the gap between reported and actual relevance stays within a specified range for each user and platform, without requiring truthful reporting.

At fixed mechanism parameters, the sample complexity of user agent's learning about the platforms' strategic behavior grows linearly with the number of platforms and falls inversely with interactions per user. 
Thus, when the benefit of accurate comparison exceeds the allowance for changing overstatement, enough feedback guarantees expected welfare at least as high as staying on the original platform. For a fixed positive margin, an interaction budget proportional to platform count suffices. A matching example shows why this cost matters: users can initially be worse off despite access to a better resource. The linear dependence cannot generally be improved for the studied mechanism.

Experiments on synthetic and Amazon dataset distinguish access to better resources from benefits users actually receive. In the synthetic market, platform credibility assessment reduces reporting losses and improves average utility over the original platform by the end of the simulated horizon. In the Amazon dataset, credibility assessment also helps correct overstatements, but observed mean utility remains below the original platform. Controlled markets connect the welfare guarantee to the feedback budget: retaining a valuable resource while adding low-value platforms requires more interactions for the guarantee utilities, and observed gains can occur before they are certified. These findings support a conditional answer to our question, not a universal benefit from competition. 

\section{Related Work}
\label{sec:related_work}

\paragraph{LLM-based recommendation and strategic presentation.}
Recommendation agents combine planning, tools, and knowledge~\citep{wang2024recmind}, support personal instructions~\citep{xu2025iagent}, and motivate user-centered services~\citep{zhang2026usercentric} and cross-platform personalization~\citep{lin2026usergoverned}. Magentic Marketplace studies consumer assistants interacting with competing service agents under different search and manipulation conditions~\citep{bansal2025magentic}. Presentation affects generated-answer visibility~\citep{aggarwal2024geo}, supplier promotion~\citep{kumar2024manipulating,nestaas2025adversarial}, recommendation rankings~\citep{filandrianos2025bias}, and shopping choices~\citep{allouah2025buying}; gains can diminish as adoption spreads~\citep{puerto2025cseo}. These effects vary across strategies and models~\citep{filandrianos2025bias}, rather than establishing a universally monotone response to exaggeration. We separate presentation from resource value to study welfare under repeated interaction and personal feedback.

\paragraph{Strategic competition and information economics.}
Strategic recommendation studies content-provider mediation~\citep{benporat2018game}, top-$K$ price of anarchy~\citep{yao2023topk}, and welfare-aligned rewards~\citep{yao2023rethinking}. Learning-platform competition need not align with user utility~\citep{jagadeesan2023competition}. Information economics examines acquisition costs~\citep{stigler1961economics}, search diversion~\citep{hagiu2011intermediaries}, and strategic communication~\citep{crawford1982strategic}. In Bayesian persuasion, \citet{collina2026emergent} identify utility and information conditions for competing providers' equilibrium outcomes to approach aligned-provider welfare. Strategic bandits address dynamic context misreports using post-selection rewards~\citep{buening2024strategic} and screen strategic click rates using post-click feedback~\citep{buening2024clickbait}, obtaining equilibrium guarantees through screening or exclusion. We fix resource-generation rules and analyze reports under softmax choice and platform credibility assessment; Appendix~\ref{app:mechanism_comparison} compares mechanisms in an overlapping submodel with the same observable feedback.

\paragraph{Credibility assessment and consumption feedback.}
Consumption provides quality information for experience goods~\citep{nelson1970information}; reputation links future returns to quality incentives~\citep{shapiro1983premiums}, while online feedback aggregates transaction experience~\citep{dellarocas2003digitization}. The Influence Limiter uses target-specific reputations to constrain raters' influence on predictions~\citep{resnick2007influence}. We instead use each user's own record to correct reports about exogenous resources, guiding selection using consumed-item feedback only. Our proof bounds welfare distortion through consumption events without requiring minimum exposure. Policy-regret analyses account for how deviations affect future feedback~\citep{arora2018policy}; our welfare guarantees instead apply under a bounded-gap condition.

\section{A Model of User-Centered Agentic Recommendation}
\label{sec:model}

\begin{figure}[!t]
 \centering
 \includegraphics[width=\linewidth]{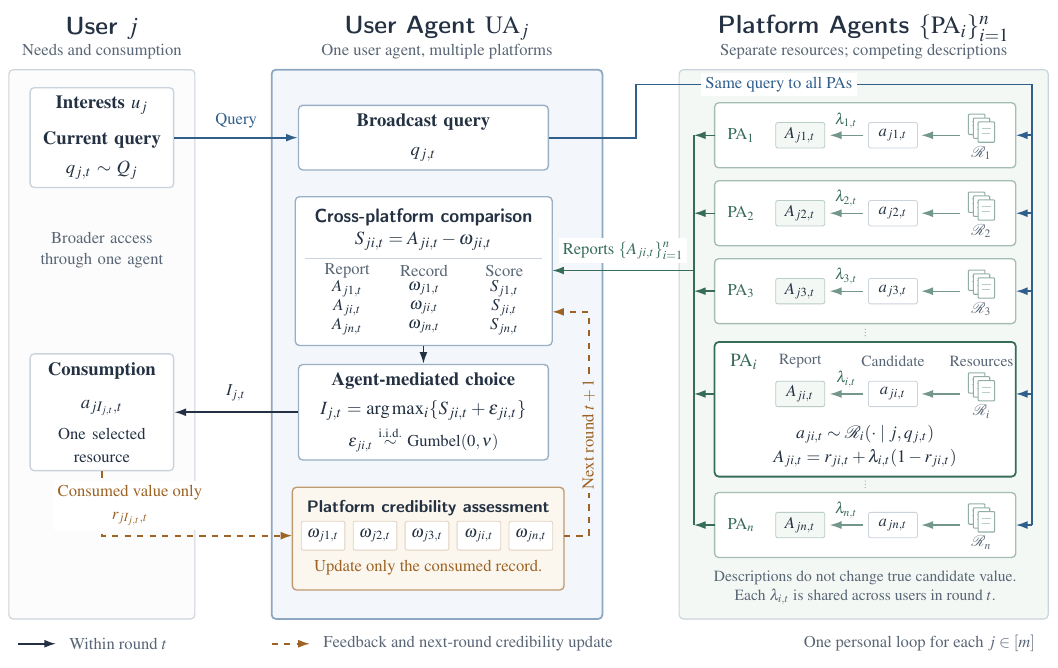}
 \caption{Interaction for user $j$ in round $t$. The UA compares recommendations from multiple PAs; after consumption, only its credibility record for the selected platform is updated.}
 \label{fig:framework_overview}
\end{figure}

We study $m$ user agents (UAs) and $n$ platform agents (PAs) interacting over $T$ rounds with a fixed relevance function $r$. Figure~\ref{fig:framework_overview} illustrates the interaction and feedback.

\paragraph{Users and platforms.}
User $j$ and platform $i$ have fixed interest centers $u_j,c_i\in\mathbb R^d$, respectively.\footnote{Centers describe generation before normalization or retrieval; Appendix~\ref{app:platform_policy} specifies the resource and information assumptions.} In round $t$, the user draws $q_{j,t}\sim Q_j$, and the platform retrieves $a_{ji,t}\sim\mathcal R_i(\cdot\mid j,q_{j,t})$ from its private resources. The fixed rule $\mathcal R_i$ may deterministically select the best local query match. Retrieved items can differ across users and rounds, but their attributes remain fixed.

\paragraph{Relevance function.}
The fixed function $r:\mathbb R^d\times\mathbb R^d\to[0,1]$ gives consumption value $r_{ji,t}=r(q_{j,t},a_{ji,t})$. This value varies with the query and retrieved candidate; expectations average over their generation rules.
A platform can make an unchanged resource appear more relevant through its description. To isolate this presentation effect without modeling the text itself, platform $i$ chooses a description intensity $\lambda_{i,t}\in[0,1]$, shared across users in round $t$. We model the resulting \emph{reported relevance} by the bounded interpolation
\begin{equation}
 A_{ji,t}=r_{ji,t}+\lambda_{i,t}(1-r_{ji,t}).
 \label{eq:claims}
\end{equation}
The intensity fills a fraction $\lambda_{i,t}$ of the remaining gap to $1$: zero is truthful and one reports maximal relevance, without changing $r_{ji,t}$ or incurring a direct cost. Equation~\eqref{eq:claims} models platform behavior for analysis; neither this map nor current intensities are supplied to the UA. It observes reports before choice and only the consumed candidate's true value afterward.

\paragraph{Recommendation policy.}
For each query, the UA compares candidates from all platforms. It uses \emph{platform credibility assessment} to account for inflated reports: a user-specific correction $\omega_{ji,t}$ summarizes overstatement observed in past consumption, giving score $S_{ji,t}=A_{ji,t}-\omega_{ji,t}$. A larger correction discounts the report more heavily. Each UA maintains its own records, so different users may assess the same platform differently.

To capture choice variability at the same scores, we use the random-utility logit model~\citep{mcfadden1974conditional,yao2023topk} with fresh i.i.d.\ shocks $\varepsilon_{ji,t}\sim\mathrm{Gumbel}(0,\nu)$, independent of the pre-choice state:
\begin{equation}
 I_{j,t}=\arg\max_{i\in[n]}\{S_{ji,t}+\varepsilon_{ji,t}\}.
 \label{eq:user_choice}
\end{equation}
Gumbel noise yields tractable selection probabilities, with a larger scale $\nu>0$ making choices less sensitive to score differences.

\paragraph{User utility.}
We evaluate recommendations by the true value of the consumed resource, not its reported relevance. Conditional on the pre-choice state, user $j$'s expected consumption utility in round $t$ is
\begin{equation}
 U_{j,t}:=\mathbb E_{\boldsymbol\varepsilon_{j,t}}
 \!\left[r_{jI_{j,t},t}\right]
 =\sum_{i=1}^n
 \Pr_{\boldsymbol\varepsilon_{j,t}}(I_{j,t}=i)\,r_{ji,t}.
 \label{eq:user_utility}
\end{equation}
The expectation holds current candidates, reports, and credibility corrections fixed; the shocks affect choice but do not enter consumption utility.

\paragraph{Platform utility.}
We measure platform payoff by value-weighted engagement, reflecting both user selection and delivered value. Under policies $\boldsymbol\pi=(\pi_1,\ldots,\pi_n)$, platform $i$'s round payoff and cumulative objective are
\begin{equation}
 u_{i,t}=\sum_{j=1}^m\mathbf1\{I_{j,t}=i\}r_{ji,t},
 \qquad J_i^T(\boldsymbol\pi)
 =\mathbb E^{\boldsymbol\pi}\!\left[\sum_{t=1}^T u_{i,t}\right].
 \label{eq:platform_payoff}
\end{equation}
Policies may adapt to locally available history, queries, and candidates; Appendix~\ref{app:platform_policy} formalizes their information constraints.

\paragraph{Platform credibility assessment.}
Records start at $\omega_{ji,1}=0$. At round end, the UA observes the consumed candidate's exact value and updates only that platform's record by exponential averaging:
\begin{equation}
 \omega_{ji,t+1}
 =\omega_{ji,t}
 +\eta\mathbf1\{I_{j,t}=i\}(A_{ji,t}-r_{ji,t}-\omega_{ji,t}),
 \qquad 0<\eta\le1.
 \label{eq:reputation}
\end{equation}
The rate $\eta$ weights each newly verified discrepancy; unselected records neither change nor decay. The credibility correction summarizes past overstatement rather than the latent intensity $\lambda_{i,t}$, and can lag changes in descriptions, queries, and retrieved items.

\begin{definition}[Actual, calibrated, and original-system utility]
\label{def:welfare}
Let $\iota_j\in[n]$ be user $j$'s original platform, fixed before outcomes and retained in the new market. Let $I^0_{j,t}$ denote the choice in~\eqref{eq:user_choice} with scores replaced by true relevance. The three average utilities are
\begin{equation}
 \begin{aligned}
 U_T(n)&=\frac1{mT}\sum_{j=1}^m\sum_{t=1}^T
       \mathbb E[U_{j,t}],\\
 U_T^0(n)&=\frac1{mT}\sum_{j=1}^m\sum_{t=1}^T
       \mathbb E[r_{jI^0_{j,t},t}],\\
 B_T&=\frac1{mT}\sum_{j=1}^m\sum_{t=1}^T
       \mathbb E[r_{j\iota_j,t}].
 \end{aligned}
 \label{eq:welfare_objects}
\end{equation}
Expectations include resource, policy, and choice randomness. Calibration uses the same candidates and noise scale, solely as an analytical reference; it provides no additional feedback to the UA. The original system retains the same candidate-generation rule but restricts each user to $\iota_j$.
\end{definition}

Our research question is: \emph{When does platform competition improve user utility over using a single fixed platform, so that $U_T(n)\ge B_T$, and how does this depend on market size, interaction horizon, and credibility updates?}

\section{Welfare Analysis}
\label{sec:welfare}
\label{sec:guarantees}

Cross-platform access can improve available resources, but strategic reports can distort choice. Using Definition~\ref{def:welfare}, write
\begin{equation}
 \mathsf{Gain}_T(n):=U_T^0(n)-B_T,\qquad
 \mathsf{Distortion}_T(n):=U_T^0(n)-U_T(n).
 \label{eq:gain_distortion}
\end{equation}
The change relative to the original platform decomposes as
\begin{equation}
 \Delta U_T(n):=U_T(n)-B_T
 =\mathsf{Gain}_T(n)-\mathsf{Distortion}_T(n).
 \label{eq:intro_decomposition}
\end{equation}
$\mathsf{Gain}_T$ measures the benefit of calibrated choice, whereas $\mathsf{Distortion}_T$ measures the effect of reporting and credibility correction; both are signed. We examine resource opportunities and competitive distortion, then bound the loss using consumption feedback.

\subsection{Resource opportunities and competitive distortion}
\label{sec:resource_opportunities}

Let $M_{j,t}(n):=\max_{i\le n}r_{ji,t}$ be the best candidate value for user $j$ in round $t$. Compare nested markets on the same query and resource realization, retaining incumbent candidates when a platform enters. Abbreviating $M_n=M_{j,t}(n)$ and $x=r_{j,n+1,t}$ gives the marginal resource opportunity
\begin{equation}
 M_{n+1}-M_n=(x-M_n)_+.
 \label{eq:frontier_increment}
\end{equation}
An entrant is valuable insofar as it supplies what the existing market lacks. For a fixed entrant and resource realization, enlarging the incumbent pool weakly reduces this opportunity, without requiring independence or identical distributions.

Appendix~\ref{app:frontier_rates} derives distributional expressions under conditional independence and growth rates under additional common-upper-tail conditions. These quantify available resources, not consumption utility.

\paragraph{From availability to consumption.}
In the calibrated benchmark, choices use true relevance but can still select a lower-value resource because of randomness. Hence
\begin{equation}
 \begin{aligned}
 \mathsf{Gain}_T(n)
 &=\frac1{mT}\sum_{j,t}\mathbb E[M_{j,t}(n)-r_{j\iota_j,t}]\\
 &\quad-\frac1{mT}\sum_{j,t}\mathbb E[M_{j,t}(n)-r_{jI^0_{j,t},t}].
 \end{aligned}
 \label{eq:frontier_gain_bridge}
\end{equation}
The first term is the gain over the original platform if the user always consumes the best available resource. The second subtracts the value lost because random choice may select a worse candidate, even with accurate scores. Better available resources therefore need not translate into higher utility.

\paragraph{Competitive reporting can erase the comparison value.}
\label{sec:competition_risk}
Even the value attainable under calibrated choice can be lost through strategic reporting. Disable credibility updates and keep $\omega_{ji,t}=0$. Let $\boldsymbol\pi^{\max}$ denote the profile in which every platform always chooses the shared intensity $\lambda_{i,t}=1$.

\begin{proposition}[Competition without credibility assessment]
\label{prop:no_reputation_equilibrium}
Under the platform payoff in~\eqref{eq:platform_payoff}, $\boldsymbol\pi^{\max}$ is a weak Nash equilibrium of the mechanism without credibility assessment over admissible finite-horizon policies. Its welfare is
\begin{equation}
 U_T^{\mathrm{off}}(\boldsymbol\pi^{\max})
 =\frac1{mT}\sum_{j,t}\mathbb E\!\left[\frac1n\sum_{i=1}^n r_{ji,t}\right].
 \label{eq:no_reputation_welfare}
\end{equation}
\end{proposition}

Against competitors reporting $1$, no unilateral policy achieves a selection probability above $1/n$. Identical reports induce uniform choice, discarding the premium from favoring better candidates (Appendix~\ref{app:off_equilibrium_proof}). This equilibrium risk motivates credibility assessment; the proposition does not assert convergence of platform learning.

\subsection{Welfare protection from consumption feedback}
\label{sec:transferable_feedback}

Only consumed resources are verified, so infrequently selected platforms can retain inaccurate records. We bound the loss from actual consumption rather than require every record to become accurate. The report--value gap $d_{ji,t}:=A_{ji,t}-r_{ji,t}=\lambda_{i,t}(1-r_{ji,t})$ changes with both the platform's action and the resource. The next condition bounds its range.

\begin{assumption}[Bounded gap amplitude]
\label{ass:amplitude}
There is a deterministic $\delta\in[0,1]$ such that, almost surely over the induced trajectories,
\begin{equation}
 \max_{s,t\le T}|d_{ji,s}-d_{ji,t}|\le\delta
 \qquad\text{for all }j\in[m],\ i\in[n].
 \label{eq:amplitude}
\end{equation}
\end{assumption}
The condition restricts variation rather than the level of overstatement, allowing large persistent discrepancies without requiring convergence. A small $\delta$ is an additional condition that adaptive reporting need not satisfy. The universally valid choice $\delta=1$ may yield an uninformative bound.

The verification budget for one personal record is
\begin{equation}
 \Psi(\nu,\eta):=\sum_{k=0}^{\infty}
 \left[1-\exp\left(-\frac{(1-\eta)^k}{\nu}\right)\right]
 \le\frac1{\nu\eta}.
 \label{eq:nonlinear_budget_series}
\end{equation}
The index $k$ counts previous consumptions of that platform, not elapsed rounds. At $\eta=1$, the first factor $(1-\eta)^0$ is interpreted as one, giving $\Psi(\nu,1)=1-e^{-1/\nu}$.

\begin{theorem}[Finite-horizon welfare guarantee]
\label{thm:amplitude}
Under Assumption~\ref{ass:amplitude}, the mechanism of Section~\ref{sec:model} satisfies
\begin{equation}
 \boxed{
 U_T(n)-B_T
 \ge\mathsf{Gain}_T(n)-\min\left\{1,\frac{n\Psi(\nu,\eta)}T+\frac{\delta}{2\nu}\right\}.}
 \label{eq:net_welfare_bound}
\end{equation}
The same error term also bounds the absolute distortion $|U_T^0(n)-U_T(n)|$.
\end{theorem}

The verification cost $n\Psi(\nu,\eta)/T$ decreases with accumulated consumption, while $\delta/(2\nu)$ accounts for mismatch from changing gaps that feedback need not eliminate. The guarantee covers every admissible, possibly adaptive profile satisfying the amplitude condition, without requiring equilibrium or truthful reporting. Corollary~\ref{cor:linearized_welfare} gives the simpler transient bound $n/(\nu\eta T)$.

\paragraph{Why consumption feedback suffices.}
\label{sec:proof_highlights}
After separating gap variation, score errors are weighted by the \emph{actual} selection probabilities. In expectation, this charges loss to consumed resources, precisely those whose records update. After $k$ previous consumptions, a pair's error beyond the gap range is at most $(1-\eta)^k$, with charge at most $1-\exp(-(1-\eta)^k/\nu)$. Its total charge is bounded by $\Psi(\nu,\eta)$; summing over $mn$ records and averaging over $mT$ interactions yields $n\Psi(\nu,\eta)/T$. No minimum exposure or unconsumed-value feedback is needed (Appendix~\ref{app:welfare_proofs}).

\paragraph{When the protected gain is positive.}
Theorem~\ref{thm:amplitude} answers the welfare question whenever calibrated gain covers the distortion budget. Let $G$ be any deterministic lower bound on $\mathsf{Gain}_T(n)$ for the same market, temperature, and horizon.

\begin{corollary}[A sufficient budget for user benefit]
\label{cor:no_drop}
\label{sec:welfare_improvement}
Under Assumption~\ref{ass:amplitude}, if $\mathsf{Gain}_T(n)\ge G>\delta/(2\nu)$, then $U_T(n)\ge B_T$ whenever
\begin{equation}
 T\ge\frac{n\Psi(\nu,\eta)}{G-\delta/(2\nu)}.
 \label{eq:sufficient_budget}
\end{equation}
\end{corollary}

For a fixed positive margin $G-\delta/(2\nu)$, the sufficient interaction budget grows linearly with market size; a smaller margin requires proportionally more interactions. Both the gain certificate and the amplitude condition must cover the evaluated horizon.

Appendix~\ref{app:market_certificates} derives explicit gain bounds under resource-separation or heterogeneous-support conditions, instantiated in Appendix~\ref{app:controlled_certificate_experiment}. These expose a temperature tradeoff: less random choice exploits resource differences more strongly but increases sensitivity to gap variation. The reporting restrictions remain imposed conditions, not consequences of payoff maximization.

\subsection{The cost of personalized verification}
\label{sec:verification_cost}

The $n/T$ dependence reflects separate user--platform records: each additional platform creates another record, but a user supplies only $T$ verifications. More users add records and feedback proportionally, so the ratio does not become $n/(mT)$. At $\eta=1$, $\Psi(\nu,1)\le1$ and the transient bound is $n/T$ even at low temperature.

The following cold-start market shows this cost even with constant values and gaps ($\delta=0$). It uses a prescribed maximal-reporting profile, not an equilibrium under credibility assessment.

\begin{samepage}
\begin{proposition}[Cold-start loss under market expansion]
\label{thm:cold_start}
Fix $c\in(0,1)$ and $n\ge4$. For one user, let the original platform have constant value $c$, one additional platform have value $1$, and the remaining $n-2$ platforms have value $0$. All reports equal $1$, $\eta=1$, and all credibility corrections start at zero. For every $\nu>0$, at $T=\lfloor n/2\rfloor$,
\begin{equation}
 U_T(n)\le\frac{2(1+c)}n,
 \qquad n>\frac{2(1+c)}c\ \Longrightarrow\ U_T(n)<B_T=c.
 \label{eq:cold_start_rate}
\end{equation}
The best available value nevertheless exceeds the original platform's value by the constant $1-c$.
\end{proposition}
\end{samepage}

Unverified reports continue attracting consumption even after the best resource is found: initial utility can be $O(1/n)$ despite a constant resource advantage (Appendix~\ref{app:verification_proofs}).

At fixed temperature, this family also dilutes calibrated choice. With $T_n=\lfloor n/2\rfloor$ and $\nu_n=1/(2\log n)$ for both choice rules, however, $U_{T_n}^0(n)\to1$ while $U_{T_n}(n)\le2(1+c)/n\to0$ (Corollary~\ref{cor:cold_start_temperature}).

In this same family, $\delta=0$ and $\Psi(\nu_n,1)=1-n^{-2}$. Theorem~\ref{thm:amplitude} also guarantees positive net utility at $T\ge Cn$, for every fixed $C>1/(1-c)$ and all sufficiently large $n$ (Corollary~\ref{cor:cold_start_matching_budget}). Thus a linear budget is enough and its order cannot be reduced for this mechanism and reporting family. This does not establish optimality across UA mechanisms or matching $\eta,\nu,\delta$ dependencies.

\section{Experiments}
\label{sec:experiments}

We evaluate access gains, reporting losses and recovery through personal consumption feedback.

\paragraph{Comparisons and protocol.}
\label{sec:experiment_setup}
We compare the original fixed platform and cross-platform choice with or without platform credibility assessment. Evaluator-only frontiers and calibrated choice at the same temperature separate availability from selection; neither supplies UA feedback. Net utility relative to the original platform decomposes into calibrated gain and distortion as in~\eqref{eq:intro_decomposition}.

We use 128 users, 4,096 rounds, 12 seeds, and nested markets $n\in\{2,4,8,16,32\}$. Unless stated otherwise, $\nu=0.1$ and $\eta=0.2$. Platform behavior spans fixed heterogeneous intensities, maximal reporting, and Exp3 learning over five intensities from zero to one~\citep{auer2002nonstochastic}. Original platforms are assigned by profile--catalog affinity among the first two catalogs before queries are drawn and retained at every width.

Comparisons reuse resources and original assignments within seeds. PA actions are held fixed in fixed-policy comparisons; learning PAs adapt to separate feedback histories. We report conditional expected consumption utility, with sampled choices generating feedback, and pointwise 95\% Student-$t$ intervals over seeds (Appendix~\ref{app:experiment_protocol}).

\subsection{Synthetic markets}
\label{sec:synthetic_results}

\paragraph{Setup.}
Users and platforms have heterogeneous interest centers in eight dimensions. Each platform has 24 private items drawn around its center. For each query drawn around a user's center, the platform samples four local items and submits the best query match. True value is cosine similarity rescaled to $[0,1]$ (Appendix~\ref{app:synthetic_details}).

\begin{figure}[!htbp]
 \centering
 \includegraphics[width=0.49\linewidth,trim=0bp 197.4bp 252bp 0bp,clip]{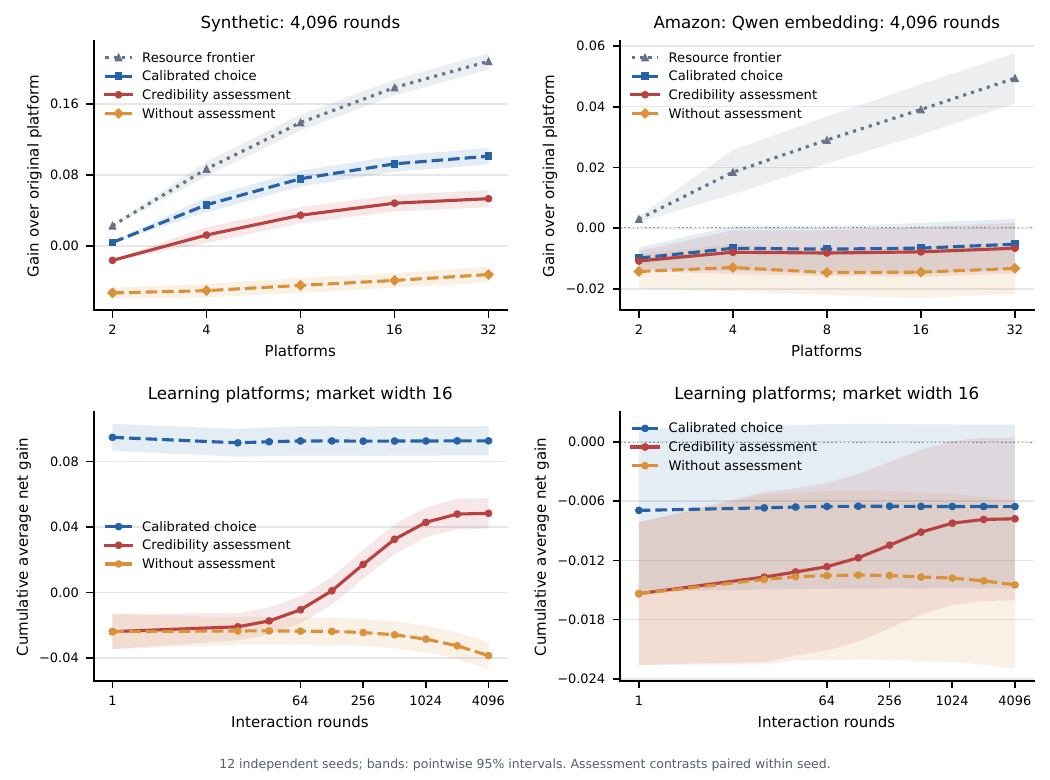}\hfill
 \includegraphics[width=0.49\linewidth,trim=0bp 19.4bp 252bp 177bp,clip]{figures/aligned_main.pdf}
 \caption{Synthetic markets under learning PAs: gains across market widths at $T=4096$ (left) and cumulative average net utility at $n=16$ (right).}
 \label{fig:synthetic_markets}
\end{figure}

\paragraph{Broader access does not remove reporting losses.}
Expanding the market improves both the frontier and calibrated choice, yet mean utility without assessment remains below the original platform at every tested width (Figure~\ref{fig:synthetic_markets}, left). These trajectories illustrate the risk in Section~\ref{sec:resource_opportunities}, without establishing convergence to its equilibrium.

\paragraph{Credibility assessment improves comparison without requiring honesty.}
Credibility assessment turns the negative mean gain positive by the final horizon (Figure~\ref{fig:synthetic_markets}, right). This does not require honesty: under fixed maximal reporting, every report remains one, yet utility exceeds the original platform (Table~\ref{tab:synthetic_reputation}). Scores $1-\omega_{ji,t}$ distinguish platforms through personal feedback even when reporting actions never improve. Useful comparison need not mean exact recovery of current item values.

Learning PAs reach net gain $0.048$ with interval $[0.039,0.058]$, despite negative cumulative gain at 64 rounds. About one third of users still lose relative to the original platform, so average benefit is not universal (Appendices~\ref{app:market_horizons}--\ref{app:amazon_sensitivity}).

Both broad-market environments fall outside the useful gap-amplitude guarantee at the tested temperatures (Appendix~\ref{app:experiment_theory_link}). Their width sweeps change resources and PA learning as well as verification, so the results describe empirical behavior rather than isolate an $n/T$ effect.

\paragraph{Market size and the guaranteed benefit region.}
\label{sec:certificate_experiment}
A separate controlled market tests Theorem~\ref{thm:amplitude} with an original platform, a preferred platform and weaker alternatives. We vary market width, preferred quality and reporting-range width $a$. PAs draw intensities independently in $[1-a,1]$ each round, without adapting to UA actions. Certificates use analytical supports, not observed extrema (Appendix~\ref{app:certificate_slices}).

Unlike broad-market expansion, weaker entrants here do not improve the best resource. Calibrated gain barely changes, but distortion rises from $0.005$ to $0.056$ between 4 and 32 platforms (Table~\ref{tab:certificate_slices}). Changes in calibrated choice also reduce net gain, so verification costs are not isolated.

Widening the reporting range can remove the positive certificate with little observed net-utility change: the theorem covers all admissible policies satisfying the range condition, not just this random policy. Widening $[1-a,1]$ also lowers mean intensity, so this is not an isolated variation effect. In a separate test with restricted Exp3, the first saved positive net-gain interval occurs at 128 rounds, versus 1,465 certified by Corollary~\ref{cor:no_drop}. The certificate gives a guaranteed region, not the empirical benefit boundary (Appendix~\ref{app:controlled_certificate_experiment}).

\subsection{Amazon-grounded markets}
\label{sec:amazon_results}

\paragraph{Setup.}
We construct virtual markets from Amazon Reviews 2023 Musical Instruments metadata and positive training interactions~\citep{hou2026bridging}. A single frozen Qwen3-Embedding-0.6B representation~\citep{zhang2025qwenembedding} embeds 2,048 candidate products and users' training profiles. Clustering the product vectors gives 32 private catalogs; these are virtual platforms, not observed sellers. Queries vary around each user's profile, and each platform samples four eligible products and submits the best query match. Value is cosine similarity rescaled to $[0,1]$, a semantic proxy rather than observed satisfaction (Appendix~\ref{app:amazon_data}).

\begin{figure}[!htbp]
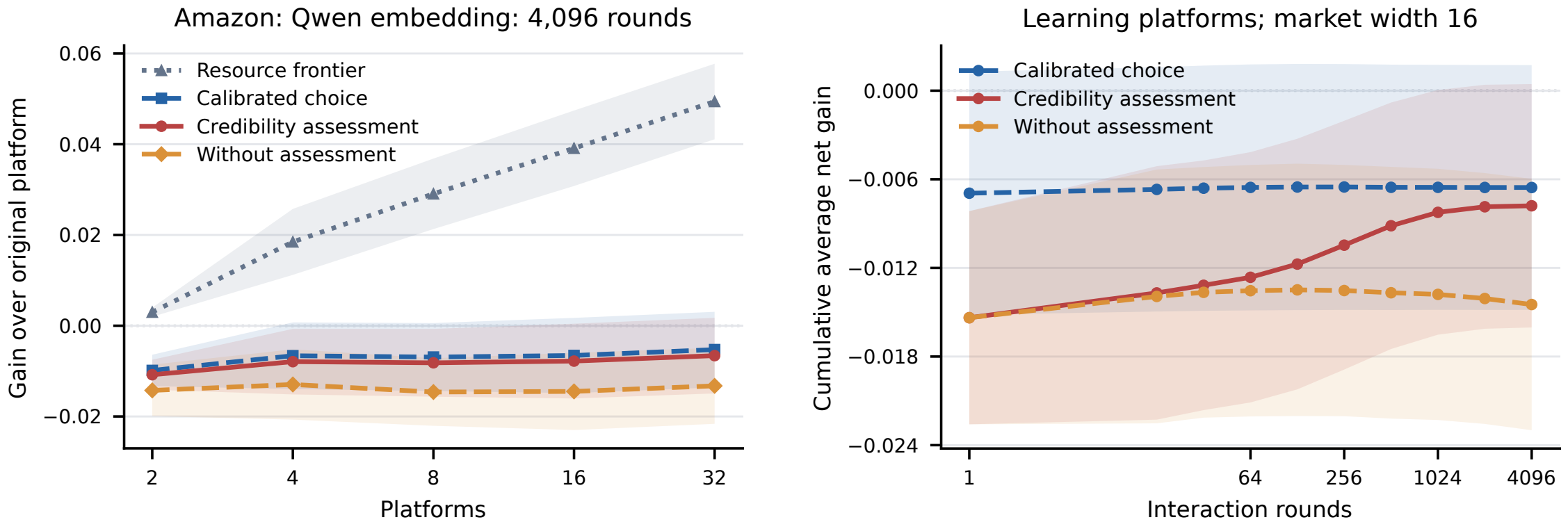

 \centering
 \includegraphics[width=0.49\linewidth,trim=252bp 197.4bp 0bp 0bp,clip]{figures/aligned_main.pdf}\hfill
 \includegraphics[width=0.49\linewidth,trim=252bp 19.4bp 0bp 177bp,clip]{figures/aligned_main.pdf}
 \caption{Amazon semantic markets under learning PAs: gains across market widths at $T=4096$ (left) and cumulative average net utility at $n=16$ (right).}
 \label{fig:amazon_markets}
\end{figure}

\paragraph{Smaller resource advantages change the value of access.}
Amazon resource frontiers improve with width, but calibrated choice has negative mean gain at every tested width under the default temperature (Figure~\ref{fig:amazon_markets}, left). This contrast precedes reporting and correction: at $n=16$, the mean frontier improvement over the original platform is $0.039$ versus $0.178$ in synthetic markets. Amazon's choice gap of $0.046$ exceeds this resource-access benefit~\eqref{eq:frontier_gain_bridge}.

\paragraph{Reducing distortion need not beat the original platform.}
Credibility assessment improves utility under all three PA conditions, including unchanged maximal reporting (Table~\ref{tab:amazon_reputation}). Under learning PAs, remaining distortion is smaller than in synthetic markets, but even zero distortion cannot overcome Amazon's negative calibrated gain. The final net-gain interval, $[-0.0160,0.0004]$, includes zero despite positive paired benefit over uncorrected reports, establishing neither average benefit nor harm relative to the original platform.

Choice depends on value differences relative to temperature. Lowering $\nu$ from $0.1$ to $0.03$ on unchanged Amazon resources makes both calibrated and actual mean gains positive, although the latter interval still includes zero (Appendix~\ref{app:amazon_sensitivity}). Under Exp3, lower temperature improves mean utility in both markets despite increasing distortion, balancing selectivity against reporting sensitivity. Net gain, not distortion alone, determines user benefit.

\paragraph{Sensitivity to the value proxy.}
Sparse observed ratings show weak or negative associations with semantic relevance in the examined cohorts. With the same retrieval process and a frozen rating-trained outcome model, credibility assessment again improves on uncorrected reports, while average net utility over the original platform becomes positive (Appendix~\ref{app:amazon_outcome_validation}). Thus the correction benefit persists across proxies, but whether users beat their original platform depends on the outcome model; neither simulation establishes actual consumer welfare.

\section{Discussion and Conclusion}
\label{sec:discussion}
\label{sec:conclusion}

User-centered agentic recommendation expands resource access while leaving personal feedback limited. Consumption-event accounting bounds welfare loss under platform credibility assessment without requiring minimum exposure. Resource and reporting conditions give explicit benefit regions; a cold-start family matches the linear dependence on market size for this mechanism. Parameterized experiments also expose substantial slack: failure of the certificate does not imply failure of the market.

The guarantees concern expected average welfare under fixed resource-generation rules, scalar reports, exact consumption feedback, zero-initialized records, and a bounded range of reporting discrepancies. Within this scope, the analysis quantifies how resource advantages and personal verification costs jointly determine the benefits of platform competition. Alternative-update comparisons appear in Appendix~\ref{app:matched_calibration}. The Amazon study evaluates simulated outcomes under specified value proxies rather than observed consumer welfare.

\subsection*{AI use statement}

We used generative AI tools to polish the manuscript, assist with writing experimental code, and help identify potential errors in the theoretical analysis.
The authors take responsibility for the final content of this work, including the theoretical claims, experimental implementation, and reported results.

\bibliography{references}

\clearpage
\appendix
\section{Resource Access: Proofs and Calibrated-Gain Bounds}
\label{app:access_proof}

\paragraph{Conditional choice notation.}
All conditional quantities below are evaluated at a realized pre-choice state, including the current queries, candidates, descriptions, and credibility records. This conditioning does not hold candidate values fixed across rounds. Write $d_{ji,t}:=A_{ji,t}-r_{ji,t}$ for the report--value gap and define
\begin{equation}
 e_{ji,t}:=d_{ji,t}-\omega_{ji,t},
 \qquad S_{ji,t}=r_{ji,t}+e_{ji,t}.
 \label{eq:residual}
\end{equation}
Let $\boldsymbol r_{j,t}=(r_{ji,t})_{i=1}^n$ and $\boldsymbol e_{j,t}=(e_{ji,t})_{i=1}^n$. The Gumbel choice rule gives the probability map and consumption-value operator
\begin{gather}
 p_i(\boldsymbol v):=\frac{\exp(v_i/\nu)}{\sum_{k=1}^n\exp(v_k/\nu)},
 \qquad p_{ji,t}=p_i(\boldsymbol r_{j,t}+\boldsymbol e_{j,t}),
 \label{eq:choice}\\
 \mathcal U_\nu(\boldsymbol r,\boldsymbol e)
 :=\sum_{i=1}^n p_i(\boldsymbol r+\boldsymbol e)r_i.
 \label{eq:utility_operator}
\end{gather}
Here $p_{ji,t}$ is conditional on the pre-choice state and averages only the fresh choice randomness. Actual conditional utility is $U_{j,t}=\mathcal U_\nu(\boldsymbol r_{j,t},\boldsymbol e_{j,t})$, and calibrated conditional utility is $\mathcal U_\nu(\boldsymbol r_{j,t},\boldsymbol0)$. Lemma~\ref{lem:gumbel_choice} derives the probabilities.

\subsection{Calibrated gains from candidate moments}
\label{app:gain_proof}

The frontier in Section~\ref{sec:resource_opportunities} quantifies available resources. The following optional certificate instead evaluates calibrated choice through candidate moments. For each user and round, define the mean and variance across the realized candidate values:
\begin{equation}
 \bar r_{j,t}:=\frac1n\sum_{i=1}^n r_{ji,t},
 \qquad
 V_{j,t}:=\frac1n\sum_{i=1}^n(r_{ji,t}-\bar r_{j,t})^2.
 \label{eq:resource_moments}
\end{equation}

\begin{proposition}[Gains from calibrated comparison]
\label{prop:calibrated_gain}
For the resource process in Section~\ref{sec:model} and any $\nu>0$,
\begin{equation}
 \mathsf{Gain}_T(n)\ge
 \frac1{mT}\sum_{j,t}\mathbb E\!\left[
 \bar r_{j,t}-r_{j\iota_j,t}
 +(1-e^{-1/\nu})V_{j,t}
 \right].
 \label{eq:gain_moment_bound}
\end{equation}
No independence or identical-distribution assumption on candidate values is required.
\end{proposition}

Denote the right-hand side of~\eqref{eq:gain_moment_bound} by $\underline G_n$. It is one possible certificate $G$ in Corollary~\ref{cor:no_drop}, not a requirement of that result. The first term is the change in average candidate quality relative to the original platform, and the second lower-bounds the selection premium over uniform choice. The certificate can be nonpositive even when calibrated gain is positive. Holding average-quality change and temperature fixed, a larger expected cross-platform variance raises this lower bound; it does not order the actual utilities of arbitrary markets. Expectations are taken after evaluating each query's candidate pool.

\begin{proof}
Fix a realized candidate-value vector $\boldsymbol r\in[0,1]^n$ for one user and round. By the Gumbel choice identity in Lemma~\ref{lem:gumbel_choice}, its calibrated conditional utility is
\[
 c:=\frac{\sum_{i=1}^n r_i e^{r_i/\nu}}
              {\sum_{i=1}^n e^{r_i/\nu}}.
\]
Suppress the user and round indices and write
\[
 \bar r:=\frac1n\sum_i r_i,\qquad
 V:=\frac1n\sum_i(r_i-\bar r)^2,\qquad
 b:=c-\bar r,\qquad \kappa:=1-e^{-1/\nu}.
\]
The definition of $c$ gives
\[
 \frac1n\sum_i(r_i-c)e^{(r_i-c)/\nu}=0,
 \qquad
 b=\frac1n\sum_i(r_i-c)\bigl(e^{(r_i-c)/\nu}-1\bigr).
\]
We claim that, for every $z\in[-1,1]$,
\[
 z(e^{z/\nu}-1)\ge\kappa z^2.
\]
For $z\ge0$, the inequalities $e^{z/\nu}-1\ge z/\nu$ and $1/\nu\ge\kappa$ prove the claim. For $z=-a\le0$ with $a\in[0,1]$, concavity of $a\mapsto1-e^{-a/\nu}$ implies
\[
 1-e^{-a/\nu}\ge a(1-e^{-1/\nu})=a\kappa.
\]
Multiplication by $a$ proves the same claim for negative $z$. Since $r_i,c\in[0,1]$, it applies to each $r_i-c$, giving
\begin{align*}
 b
 &\ge\frac{\kappa}{n}\sum_i(r_i-c)^2\\
 &=\kappa\bigl(V+(c-\bar r)^2\bigr)
 =\kappa(V+b^2)\ge\kappa V.
\end{align*}
Thus calibrated conditional utility is at least $\bar r+\kappa V$ for every realized vector. Apply this inequality to $\boldsymbol r_{j,t}$, subtract the same-round value $r_{j\iota_j,t}$ of the original platform, and take expectations and the average over users and rounds. Definition~\ref{def:welfare} then yields~\eqref{eq:gain_moment_bound}. The argument uses no independence, identical-distribution, or constant-value assumption: the expectation is taken after evaluating calibrated utility and the cross-platform variance at each realized vector.
\end{proof}

\subsection{Proof of Proposition~\ref{prop:no_reputation_equilibrium}}
\label{app:off_equilibrium_proof}

\begin{proof}
In the without-assessment mechanism, credibility corrections are identically zero. Fix any platform $i$ and any admissible, possibly randomized and history-dependent deviation $\pi_i'$, while every other platform uses the constant policy $\lambda_{k,t}=1$. For every pre-choice state reached under this deviation, all rivals report $1$, whereas $A_{ji,t}=r_{ji,t}+\lambda_{i,t}(1-r_{ji,t})\le1$. Fresh Gumbel choice randomness therefore gives the conditional probability
\[
 p_{ji,t}'
 =\frac{e^{A_{ji,t}/\nu}}
        {e^{A_{ji,t}/\nu}+(n-1)e^{1/\nu}}
 \le\frac1n.
\]
As $r_{ji,t}\ge0$, the deviating platform's conditional expected round payoff is at most
\[
 \sum_{j=1}^m r_{ji,t}p_{ji,t}'
 \le\frac1n\sum_{j=1}^m r_{ji,t}.
\]
This holds for every shared intensity chosen by the deviation, including choices based on current candidates and all admissible past feedback. The joint query and candidate law is independent of description policies, so the resource expectations on the right are unchanged by the deviation. Summing over rounds and taking expectations yields
\[
 J_i^{T,\mathrm{off}}(\pi_i',\boldsymbol\pi_{-i}^{\max})
 \le\frac1n\sum_{j=1}^m\sum_{t=1}^T\mathbb E[r_{ji,t}]
 =J_i^{T,\mathrm{off}}(\boldsymbol\pi^{\max}).
\]
The equality holds because all scores equal $1$ under $\boldsymbol\pi^{\max}$, so each platform is selected with conditional probability $1/n$. Since the comparison covers every admissible full-horizon unilateral deviation, $\boldsymbol\pi^{\max}$ is a weak Nash equilibrium. The proof permits dependencies across platforms, users, and rounds; it does not require a fixed realized resource vector.

Under the same all-maximal policy profile, each user's conditional expected consumption value is $n^{-1}\sum_i r_{ji,t}=\bar r_{j,t}$. Averaging gives~\eqref{eq:no_reputation_welfare}. Consequently, the difference between calibrated utility and this equilibrium's utility is exactly the average selection premium over uniform choice. Applying the pointwise inequality proved above gives
\begin{equation}
 U_T^0(n)-U_T^{\mathrm{off}}(\boldsymbol\pi^{\max})
 \ge \frac{1-e^{-1/\nu}}{mT}\sum_{j,t}\mathbb E[V_{j,t}].
 \label{eq:no_reputation_premium}
\end{equation}
This quantifies the comparison value lost at the without-assessment equilibrium. The result does not assert dominance against arbitrary history-responsive rivals, uniqueness, or convergence of a learning process.
\end{proof}

\subsection{Marginal value of an additional candidate}

\begin{proposition}[The value of an additional candidate]
\label{prop:candidate_increment}
Fix a user, a round, and a temperature $\nu>0$. For values $r_1,\ldots,r_n$, let $Z_n=\sum_{i=1}^n e^{r_i/\nu}$, $u_n^0=\mathcal U_\nu(\boldsymbol r,\boldsymbol0)$, and $M_n=\max_{i\le n}r_i$. Adding a candidate of value $x\in[0,1]$ changes calibrated utility by
\begin{equation}
 u_{n+1}^0-u_n^0
 =\frac{e^{x/\nu}}{Z_n+e^{x/\nu}}(x-u_n^0),
 \label{eq:calibrated_increment}
\end{equation}
The corresponding frontier increment is~\eqref{eq:frontier_increment}.
\end{proposition}

\begin{proof}
For the calibrated choice rule, adding value $x$ gives
\[
 u_{n+1}^0=\frac{Z_nu_n^0+xe^{x/\nu}}{Z_n+e^{x/\nu}}.
\]
Subtracting $u_n^0$ proves Equation~\eqref{eq:calibrated_increment}. Its coefficient is strictly positive, so the sign of the increment is the sign of $x-u_n^0$. The frontier identity follows directly from
$M_{n+1}=\max\{M_n,x\}$.
\end{proof}

\subsection{Resource-frontier results and proofs}
\label{app:frontier_rates}

To quantify the opportunity for heterogeneous platforms, write $F_{ji,t}(x):=\Pr(r_{ji,t}\le x\mid q_{j,t})$.

\begin{proposition}[Marginal resource opportunity]
\label{prop:frontier}
Suppose candidate values are conditionally independent across platforms given $q_{j,t}$, with conditional CDFs $F_{ji,t}$ that need not coincide. Then
\begin{equation}
  \mathbb E\!\left[M_{j,t}(n+1)-M_{j,t}(n)\mid q_{j,t}\right]
  =\int_0^1\prod_{i=1}^{n}F_{ji,t}(x)
    \bigl[1-F_{j,n+1,t}(x)\bigr]\,dx.
  \label{eq:frontier_cdf}
\end{equation}
\end{proposition}

At each quality threshold, the integrand is the probability that no incumbent exceeds it while the entrant does. Thus the marginal value of a platform depends on its resources relative to the entire incumbent pool, not on a representative platform distribution.

An explicit rate in market width follows under an additional common-upper-tail benchmark. Suppose the conditional distributions share right endpoint $1$ and, for almost every query, satisfy
\begin{equation}
  c_-x^\alpha\le 1-F_{ji,t}(1-x)\le c_+x^\alpha,
  \qquad 0<x\le x_0,
  \label{eq:frontier_tail}
\end{equation}
where $c_-,c_+,\alpha>0$ and $x_0\in(0,1]$ are uniform over users, rounds, queries, and a fixed nested sequence of platforms. Platforms may remain heterogeneous within these bounds.

\begin{corollary}[Growth under a common upper tail]
\label{cor:frontier_rates}
Under the conditional independence in Proposition~\ref{prop:frontier} and the uniform tail condition~\eqref{eq:frontier_tail}, as $n\to\infty$,
\begin{align}
  \mathbb E[1-M_{j,t}(n)\mid q_{j,t}]
    &=\Theta(n^{-1/\alpha}),
  \label{eq:frontier_rates}\\
  \mathbb E[M_{j,t}(n+1)-M_{j,t}(n)\mid q_{j,t}]
    &=\Theta(n^{-1-1/\alpha}).
  \label{eq:conditional_frontier_rates}
\end{align}
The constants are uniform, so the same orders hold after averaging over users, rounds, and queries.
\end{corollary}

The exponent $\alpha$ describes the scarcity of near-optimal resources: a larger exponent gives a slower approach to the upper endpoint. These rates quantify diminishing opportunities at the market scale, although different heterogeneous entrants need not have successively smaller increments.

Proposition~\ref{prop:frontier} allows discrete or continuous candidate values. Corollary~\ref{cor:frontier_rates} additionally requires the common-endpoint, uniform upper-tail benchmark. Given a query and a realized finite catalog, candidate values have finite support and cannot satisfy~\eqref{eq:frontier_tail} with $\alpha>0$: a maximum below $1$ violates the positive lower tail bound, whereas an atom at $1$ violates the upper bound as $x\downarrow0$. Thus these rates do not certify the fixed-catalog experiments.

The upper-tail benchmark is nevertheless compatible with immutable items when its expectation includes random resource initialization. For example, fix a query and independently draw one item value $X_i\sim\mathrm{Unif}[0,1]$ per platform at initialization, retaining that item forever. With $r(q,a)=\operatorname{clip}_{[0,1]}(a)$ and $a_{ji,t}=X_i$, the conditional CDF before conditioning on the realized items is continuous uniform in every round. No item changes across rounds. More generally, any such interpretation must verify conditional independence and the tail bounds under the same conditioning. The rates describe the opportunity term in~\eqref{eq:frontier_gain_bridge}, not a growth law for calibrated or actual utility.

\begin{proof}
Fix a user, round, and query, and take all probabilities and expectations in this proof conditional on that query. Write $R_i=r_{ji,t}$, $M_n=\max_{i\le n}R_i$, and $F_i(x)=\Pr(R_i\le x\mid q_{j,t})$. No independence across users or rounds is required.

\paragraph{Exact marginal opportunity.}
For $X=R_{n+1}$, the pointwise identity
\[
  (X-M_n)_+=\int_0^1\mathbf 1\{M_n\le x<X\}\,dx
\]
holds because all values lie in $[0,1]$. The same-entrant comparison in Section~\ref{sec:resource_opportunities} is also pointwise: under a common resource realization, nested nonempty incumbent sets $\mathcal S\subseteq\mathcal T$ not containing entrant $k$ satisfy
\[
 (R_k-\max_{i\in\mathcal T}R_i)_+
 \le (R_k-\max_{i\in\mathcal S}R_i)_+.
\]
Taking conditional expectations preserves the inequality without an independence assumption. This compares the same entrant against two incumbent sets; it does not compare different entrants along a heterogeneous arrival sequence.

Tonelli's theorem and conditional independence across platforms give
\begin{align*}
  \mathbb E[M_{n+1}-M_n]
  &=\int_0^1\Pr(M_n\le x,\,X>x)\,dx\\
  &=\int_0^1\left[\prod_{i=1}^{n}F_i(x)\right]
       [1-F_{n+1}(x)]\,dx.
\end{align*}
This proves Equation~\eqref{eq:frontier_cdf}, including distributions with atoms.

\paragraph{Distance to the upper endpoint.}
Now impose Equation~\eqref{eq:frontier_tail} with constants uniform over the nested platform sequence and all conditioning queries. Set
\[
  b=\min\{x_0,(2c_+)^{-1/\alpha}\},
  \qquad \rho=1-c_-b^\alpha\in[1/2,1).
\]
For $Y_n=1-M_n$ and $0<x\le b$, conditional independence yields
\begin{equation}
  (1-c_+x^\alpha)^n
  \le \Pr(Y_n\ge x)
  =\prod_{i=1}^{n}F_i(1-x)
  \le e^{-nc_-x^\alpha}.
  \label{eq:app_frontier_tail_bound}
\end{equation}
The non-strict inequality in $Y_n\ge x$ respects possible atoms; endpoints do not affect the tail integral. For $x>b$, monotonicity of the CDFs gives $\Pr(Y_n\ge x)\le\rho^n$. Hence
\begin{align*}
  \mathbb E[Y_n]
  &\le\int_0^\infty e^{-nc_-x^\alpha}\,dx+\rho^n\\
  &=\Gamma\!\left(1+\frac1\alpha\right)
       (nc_-)^{-1/\alpha}+\rho^n.
\end{align*}
For the lower bound, $c_+x^\alpha\le1/2$ on $[0,b]$, so
$\log(1-c_+x^\alpha)\ge-2c_+x^\alpha$. Restricting the integral to
$[0,bn^{-1/\alpha}]$ gives
\begin{align*}
  \mathbb E[Y_n]
  &\ge\int_0^{bn^{-1/\alpha}}e^{-2nc_+x^\alpha}\,dx\\
  &\ge b e^{-2c_+b^\alpha}n^{-1/\alpha}.
\end{align*}
The exponentially small $\rho^n$ is dominated by $n^{-1/\alpha}$ as $n\to\infty$, proving the first rate.

\paragraph{Marginal rate.}
Subtracting two $\Theta$ bounds does not determine the order of an increment. Instead, change variables in the exact marginal formula to obtain
\[
  \mathbb E[M_{n+1}-M_n]
  =\int_0^1\left[\prod_{i=1}^{n}F_i(1-x)\right]
     [1-F_{n+1}(1-x)]\,dx.
\]
On $[0,b]$, apply Equation~\eqref{eq:app_frontier_tail_bound} to the product and the two-sided tail condition to the new platform. Beyond $b$, bound the integrand by $\rho^n$. The upper bound is
\begin{align*}
  \mathbb E[M_{n+1}-M_n]
  &\le c_+\int_0^\infty x^\alpha e^{-nc_-x^\alpha}\,dx+\rho^n\\
  &=\frac{c_+}{\alpha}\Gamma\!\left(1+\frac1\alpha\right)
       (nc_-)^{-1-1/\alpha}+\rho^n.
\end{align*}
The corresponding lower bound is
\begin{align*}
  \mathbb E[M_{n+1}-M_n]
  &\ge c_-\int_0^{bn^{-1/\alpha}}
           x^\alpha e^{-2nc_+x^\alpha}\,dx\\
  &\ge\frac{c_-b^{\alpha+1}e^{-2c_+b^\alpha}}{\alpha+1}
       n^{-1-1/\alpha}.
\end{align*}
These establish the second rate. Every constant depends only on
$(c_-,c_+,x_0,\alpha)$, not on platform identity, the user, round, or query. Taking expectations over queries and averaging over users and rounds therefore preserves both orders. The argument requires conditional independence, but nowhere requires identical platform distributions.
\end{proof}

\subsection{Why frontier rates are not calibrated-gain rates}

\begin{remark}[A frontier rate need not be a utility rate]
For an illustrative i.i.d.\ continuous uniform benchmark, $\Pr(M_n\le x)=x^n$ on $[0,1]$, and hence
\[
  \mathbb E[M_n]=\int_0^1(1-x^n)\,dx=\frac n{n+1}.
\]
For any fixed $\nu>0$, the strong law of large numbers applies to the bounded variables $R_i e^{R_i/\nu}$ and $e^{R_i/\nu}$. Their sample means converge almost surely, and the denominator has a strictly positive limit. Thus
\[
  u_n^0
  \longrightarrow
  \frac{\int_0^1x e^{x/\nu}\,dx}{\int_0^1 e^{x/\nu}\,dx}
  \quad\text{almost surely}.
\]
Since $0\le u_n^0\le1$, bounded convergence also gives convergence of expectations. Evaluating the two integrals,
\[
  \int_0^1e^{x/\nu}\,dx=\nu(e^{1/\nu}-1),
  \qquad
  \int_0^1x e^{x/\nu}\,dx
    =\nu e^{1/\nu}-\nu^2(e^{1/\nu}-1),
\]
gives
\begin{equation}
 \mathbb E[M_n]=\frac n{n+1}\longrightarrow1,\qquad
 \mathbb E[u_n^0]\longrightarrow
 1-\nu+\frac1{e^{1/\nu}-1}.
 \label{eq:frontier_softmax_example}
\end{equation}
 It is strictly below $1$ because $e^{1/\nu}-1>1/\nu$. This example does not impose identical distributions on the main model: even a simple access benchmark does not justify identifying the frontier rate with a calibrated-gain rate at fixed temperature.
\end{remark}

\section{Proofs of the Consumption-Based Welfare Results}
\label{app:welfare_proofs}

We derive the choice probabilities and perturbation bounds, prove the consumption-based welfare guarantee, and give computable verification budgets and resource-based gain certificates.

\subsection{From Gumbel choice to selection probabilities}
\label{app:gumbel_choice}

\begin{lemma}[Gumbel choice probabilities]
\label{lem:gumbel_choice}
Conditional on the pre-choice state, let $S_1,\ldots,S_n$ be fixed scores and let $\varepsilon_i$ be independent $\mathrm{Gumbel}(\mu,\nu)$ shocks with common location $\mu$ and scale $\nu>0$. If $I=\arg\max_i(S_i+\varepsilon_i)$, then
\[
 \Pr(I=i)=\frac{\exp(S_i/\nu)}{\sum_k\exp(S_k/\nu)}.
\]
\end{lemma}
\begin{proof}
The Gumbel cumulative distribution and density are
\[
 F(x)=\exp\!\left[-e^{-(x-\mu)/\nu}\right],\qquad
 f(x)=\frac1\nu e^{-(x-\mu)/\nu}F(x).
\]
Ties have probability zero. Conditioning on $\varepsilon_i=x$ and using independence gives
\[
 \Pr(I=i)=\int_{-\infty}^{\infty}
 f(x)\prod_{k\ne i}F(x+S_i-S_k)\,dx.
\]
Set $y=e^{-(x-\mu)/\nu}$. Then $f(x)\,dx=-e^{-y}\,dy$, so reversing the integration limits yields
\[
 \begin{aligned}
 \Pr(I=i)
 &=\int_0^\infty
 \exp\!\left[-y\left(1+\sum_{k\ne i}e^{(S_k-S_i)/\nu}\right)\right]dy\\
 &=\frac1{\sum_k e^{(S_k-S_i)/\nu}}
 =\frac{e^{S_i/\nu}}{\sum_k e^{S_k/\nu}}.
 \end{aligned}
\]
The common location cancels. For fixed true values $r_i$, it follows that
$\mathbb E[r_I]=\sum_i\Pr(I=i)r_i$, which gives~\eqref{eq:utility_operator} upon substituting $S_i=r_i+e_i$. The expectation does not include the choice shock in the reward.
\end{proof}

\subsection{Softmax perturbations}

The following lemmas provide the two comparisons used in Theorem~\ref{thm:amplitude}. All vectors refer to one user and one round.

\begin{lemma}[Nonnegative reweighting]
\label{lem:reweight}
Let $\boldsymbol v\in\mathbb R^n$ be arbitrary base scores, $\boldsymbol z\ge0$ score increments, and $\boldsymbol r\in[0,1]^n$ rewards. Let $\boldsymbol p^0$ and $\boldsymbol p$ be the softmax distributions of $\boldsymbol v$ and $\boldsymbol v+\boldsymbol z$ at temperature $\nu>0$. Then
\begin{equation}
 |\langle\boldsymbol p-\boldsymbol p^0,\boldsymbol r\rangle|
 \le \sum_i p_i(1-e^{-z_i/\nu})
 \le \frac1\nu\sum_i p_i z_i.
 \label{eq:reweight}
\end{equation}
\end{lemma}
\begin{proof}
Put $Z=\sum_i p_i e^{-z_i/\nu}\in(0,1]$. Direct normalization gives
$p_i^0=p_i e^{-z_i/\nu}/Z$. If $Z<1$, the vector
$q_i=p_i(1-e^{-z_i/\nu})/(1-Z)$ is a probability distribution and
$\boldsymbol p=Z\boldsymbol p^0+(1-Z)\boldsymbol q$.
Both expected rewards lie in $[0,1]$, so their difference is at most $1-Z$ in absolute value. The final bound follows from $1-e^{-x}\le x$ for $x\ge0$. If $Z=1$, positivity of softmax probabilities implies $z_i=0$ for every $i$, and the distributions agree. The argument requires no relation between the base scores $\boldsymbol v$ and the true rewards $\boldsymbol r$.
\end{proof}

\begin{lemma}[Relative score perturbations]
\label{lem:oscillation}
For $\boldsymbol r\in[0,1]^n$, $\boldsymbol b\in\mathbb R^n$, and $\nu>0$, write $\operatorname{osc}(\boldsymbol x)=\max_i x_i-\min_i x_i$. Then
\begin{equation}
 |\mathcal U_\nu(\boldsymbol r,\boldsymbol b)-\mathcal U_\nu(\boldsymbol r,\boldsymbol0)|
 \le \operatorname{osc}(\boldsymbol r)
       \min\left\{1,\frac{\operatorname{osc}(\boldsymbol b)}{4\nu}\right\}.
 \label{eq:oscillation}
\end{equation}
\end{lemma}
\begin{proof}
Let $p_i^{(s)}\propto\exp((r_i+s b_i)/\nu)$ for $s\in[0,1]$. Differentiating the normalization gives
\[
 \frac{d}{ds}\sum_i p_i^{(s)}r_i
 =\frac1\nu\operatorname{Cov}_{\boldsymbol p^{(s)}}(\boldsymbol r,\boldsymbol b).
\]
A random variable supported on an interval of length $L$ has variance at most $L^2/4$: subtract the interval midpoint and bound the resulting second moment. Cauchy--Schwarz therefore bounds the absolute covariance by
$\operatorname{osc}(\boldsymbol r)\operatorname{osc}(\boldsymbol b)/4$.
Integration proves the linear bound. The other bound follows because both expected rewards lie between $\min_i r_i$ and $\max_i r_i$.
\end{proof}

\subsection{Proof of Theorem~\ref{thm:amplitude}}

We first establish the theorem's absolute distortion guarantee:
\begin{equation}
 |\mathsf{Distortion}_T(n)|
 \le\min\left\{1,\frac{n\Psi(\nu,\eta)}T+\frac{\delta}{2\nu}\right\}.
 \label{eq:nonlinear_amplitude_bound}
\end{equation}
Combining this bound with~\eqref{eq:intro_decomposition} yields~\eqref{eq:net_welfare_bound}.

\begin{proof}
Fix a trajectory on which Assumption~\ref{ass:amplitude} holds and put
$a_{ji}=\min_{s\le T}d_{ji,s}$. This path-dependent minimum is a proof aid, not an algorithmic input. If pair $(j,i)$ has been selected $N$ times before round $t$, expanding the convex-combination update \eqref{eq:reputation} gives
\begin{equation}
 a_{ji}[1-(1-\eta)^N]\le\omega_{ji,t}\le a_{ji}+\delta,
 \qquad
 -\delta\le e_{ji,t}\le\delta+a_{ji}(1-\eta)^N.
 \label{eq:amplitude_state}
\end{equation}
For $\eta=1$, the product interpretation gives $(1-\eta)^N=1$ when $N=0$ and $0$ otherwise.

Define pre-choice auxiliary variables
\[
 b_{ji,t}=\min\{e_{ji,t},\delta\},\qquad
 z_{ji,t}=(e_{ji,t}-\delta)_+.
\]
Then $e=b+z$, $b\in[-\delta,\delta]$, and $z\ge0$. Apply Lemma~\ref{lem:reweight} between scores $\boldsymbol r+\boldsymbol b+\boldsymbol z$ and $\boldsymbol r+\boldsymbol b$, and Lemma~\ref{lem:oscillation} between $\boldsymbol r+\boldsymbol b$ and $\boldsymbol r$. The first comparison is weighted by the actual selection probabilities:
\begin{equation}
 |\mathcal U_\nu(\boldsymbol r_{j,t},\boldsymbol e_{j,t})
  -\mathcal U_\nu(\boldsymbol r_{j,t},\boldsymbol0)|
 \le\sum_i p_{ji,t}(1-e^{-z_{ji,t}/\nu})+\frac{\delta}{2\nu}.
 \label{eq:amplitude_local}
\end{equation}
Let $\mathcal F_t$ be the complete pre-choice analysis state, including current values, reports, and credibility records; it is not the UA's observation. Since $\delta$ is deterministic and $z$ is determined by the current state, fresh choice randomness gives
\[
 \mathbb E[1-e^{-z_{jI_{j,t},t}/\nu}\mid\mathcal F_t]
 =\sum_i p_{ji,t}(1-e^{-z_{ji,t}/\nu}).
\]
On the $(k+1)$st selection of a pair, \eqref{eq:amplitude_state} implies $z\le a_{ji}(1-\eta)^k\le(1-\eta)^k$. Thus, pathwise,
\begin{equation}
 \sum_{\{t:I_{j,t}=i\}}(1-e^{-z_{ji,t}/\nu})
 \le\sum_{k\ge0}\left[1-\exp\left(-\frac{(1-\eta)^k}{\nu}\right)\right]
 =\Psi(\nu,\eta).
 \label{eq:nonlinear_selected_budget}
\end{equation}
Take the triangle inequality for the averaged distortion, substitute the conditional-expectation identity, sum over the $mn$ pairs, and divide by $mT$ in \eqref{eq:amplitude_local}. This proves \eqref{eq:nonlinear_amplitude_bound}, with truncation at $1$ because both utilities lie in $[0,1]$. Substitution in~\eqref{eq:intro_decomposition} proves the net-welfare bound~\eqref{eq:net_welfare_bound}. The future-dependent $a_{ji}$ enters only the pathwise upper bound, not the conditional-expectation substitution.

The welfare bound concerns expectations; the pathwise accounting step does not turn it into a realized or high-probability guarantee.
\end{proof}

\begin{corollary}[Linearized welfare bound]
\label{cor:linearized_welfare}
Under Assumption~\ref{ass:amplitude},
\begin{equation}
 |\mathsf{Distortion}_T(n)|
 \le\min\left\{1,\frac{n}{\nu\eta T}+\frac{\delta}{2\nu}\right\}.
 \label{eq:amplitude_bound}
\end{equation}
\end{corollary}
\begin{proof}
Apply $1-e^{-x}\le x$ termwise in~\eqref{eq:nonlinear_budget_series} to obtain
$\Psi(\nu,\eta)\le\nu^{-1}\sum_{k\ge0}(1-\eta)^k=1/(\nu\eta)$,
and substitute into~\eqref{eq:nonlinear_amplitude_bound}.
\end{proof}

\paragraph{Comparison with the without-assessment equilibrium.}
\label{app:reputation_off_comparison}
For the same resource market and temperature, combining~\eqref{eq:no_reputation_premium} with~\eqref{eq:nonlinear_amplitude_bound} gives, for every admissible credibility-assessment game profile satisfying Assumption~\ref{ass:amplitude},
\begin{equation}
 \begin{aligned}
 U_T(n)-U_T^{\mathrm{off}}(\boldsymbol\pi^{\max})
 &\ge \frac{1-e^{-1/\nu}}{mT}\sum_{j,t}\mathbb E[V_{j,t}]\\
 &\quad-\min\left\{1,\frac{n\Psi(\nu,\eta)}T+\frac{\delta}{2\nu}\right\}.
 \end{aligned}
 \label{eq:reputation_pooling_comparison}
\end{equation}
This follows by adding and subtracting $U_T^0(n)$. It compares the profile with the specified without-assessment equilibrium, without requiring the credibility-assessment game profile to be an equilibrium.

\subsection{Evaluating the verification budget}
\label{app:nonlinear_budget}

The budget in~\eqref{eq:nonlinear_budget_series} retains the nonlinear cost of changing a selection distribution. It is a property of the existing mechanism, not an additional algorithmic parameter.

\begin{proposition}[Budget evaluation and certification]
\label{prop:nonlinear_budget}
For every $\nu>0$, $0<\eta\le1$, and integer $K\ge1$,
\begin{equation}
 0\le\Psi(\nu,\eta)-\sum_{k=0}^{K-1}
 \left[1-\exp\left(-\frac{(1-\eta)^k}{\nu}\right)\right]
 \le\frac{(1-\eta)^K}{\nu\eta}.
 \label{eq:nonlinear_budget_tail}
\end{equation}
Under Assumption~\ref{ass:amplitude}, if a deterministic certificate $G$ satisfies $\mathsf{Gain}_T(n)\ge G>\delta/(2\nu)$ in the evaluated market and horizon, then
\begin{equation}
 T\ge\frac{n\Psi(\nu,\eta)}{G-\delta/(2\nu)}
 \quad\Longrightarrow\quad U_T(n)\ge B_T.
 \label{eq:nonlinear_sufficient_budget}
\end{equation}
\end{proposition}

\begin{proof}
Bound the omitted terms by $(1-\eta)^k/\nu$ and sum their geometric tail to obtain~\eqref{eq:nonlinear_budget_tail}. Substituting the assumed gain certificate into Theorem~\ref{thm:amplitude} and solving for $T$ proves~\eqref{eq:nonlinear_sufficient_budget}, equivalently Corollary~\ref{cor:no_drop}.
\end{proof}

At $\nu=0.1$ and $\eta=0.2$, this yields $\Psi(\nu,\eta)\approx13.405600945$, compared with $1/(\nu\eta)=50$. The refinement reduces the verification allowance but leaves the gap-amplitude requirement unchanged.

\subsection{Certificates from resource and reporting conditions}
\label{app:market_certificates}

Theorem~\ref{thm:amplitude} requires a gap-amplitude bound, while Corollary~\ref{cor:no_drop} also requires positive calibrated gain. The following certificates derive both from explicit resource and reporting conditions. They are sufficient conditions on a market, not extra observations supplied to the UA and not predictions of strategic behavior.

\begin{proposition}[Heterogeneous support certificate]
\label{prop:support_certificate}
Suppose deterministic intervals satisfy, almost surely throughout the evaluated horizon,
\[
 r_{ji,t}\in[L_{ji},H_{ji}]\subseteq[0,1],\qquad
 \lambda_{i,t}\in[\lambda_i^-,\lambda_i^+]\subseteq[0,1].
\]
The platform intensity remains shared across users and may otherwise adapt arbitrarily. Set
\begin{equation}
 \delta_{\mathrm{sup}}:=\max_{j,i}
 \left\{\lambda_i^+(1-L_{ji})-\lambda_i^-(1-H_{ji})\right\}.
 \label{eq:support_amplitude}
\end{equation}
For each user choose a reference platform $k_j$, put $L_j^{\min}=\min_i L_{ji}$, and define
\begin{equation}
 \begin{aligned}
 G_j^{\mathrm{sup}}
 &:=L_j^{\min}-H_{j\iota_j}
 +\frac{L_{jk_j}-L_j^{\min}}
 {1+\sum_{i\ne k_j}\exp((H_{ji}-L_{jk_j})/\nu)},\\
 G_{\mathrm{sup}}&:=\frac1m\sum_{j=1}^m G_j^{\mathrm{sup}}.
 \end{aligned}
 \label{eq:support_gain}
\end{equation}
Then Assumption~\ref{ass:amplitude} holds with $\delta=\delta_{\mathrm{sup}}$ and $\mathsf{Gain}_T(n)\ge G_{\mathrm{sup}}$. Hence, if $G_{\mathrm{sup}}>\delta_{\mathrm{sup}}/(2\nu)$,
\begin{equation}
 T\ge\frac{n\Psi(\nu,\eta)}{G_{\mathrm{sup}}-\delta_{\mathrm{sup}}/(2\nu)}
 \quad\Longrightarrow\quad U_T(n)\ge B_T.
 \label{eq:support_budget}
\end{equation}
\end{proposition}

\begin{proof}
The map $\lambda(1-r)$ is increasing in $\lambda$ and decreasing in $r$ on the given domain. Therefore each pair's gap stays in
\[
 [\lambda_i^-(1-H_{ji}),\ \lambda_i^+(1-L_{ji})],
\]
which proves~\eqref{eq:support_amplitude}, simultaneously for all users under a shared platform action. Conditional on any candidate realization, calibrated choice gives
\[
 p^0_{jk_j,t}
 =\frac1{1+\sum_{i\ne k_j}\exp((r_{ji,t}-r_{jk_j,t})/\nu)}
 \ge\frac1{1+\sum_{i\ne k_j}\exp((H_{ji}-L_{jk_j})/\nu)}.
\]
Every candidate has value at least $L_j^{\min}$, and candidate $k_j$ has value at least $L_{jk_j}$. Thus calibrated conditional utility minus the original value is at least
\[
 L_j^{\min}+p^0_{jk_j,t}(L_{jk_j}-L_j^{\min})-H_{j\iota_j}
 \ge G_j^{\mathrm{sup}}.
\]
The multiplier $L_{jk_j}-L_j^{\min}$ is nonnegative, so the probability lower bound preserves the inequality. Average over candidates, rounds, and users to prove~\eqref{eq:support_gain}; apply Corollary~\ref{cor:no_drop} to obtain~\eqref{eq:support_budget}.
\end{proof}

Independence, identical distributions, and deterministic candidate values are unnecessary. The certificate is valid for any reference $k_j$, although choosing a platform with a high lower endpoint makes it useful. In particular, distinct users may have distinct reference platforms. The intervals must cover the induced distributions, not merely the values observed in one simulation. For example, if reporting lies in $[1-a,1]$ and a pair's value lies in $[v-w,v+w]\subseteq[0,1]$, its gap range is at most
\begin{equation}
 2w+a(1-v-w).
 \label{eq:support_symmetric_family}
\end{equation}
This expression separates resource spread from allowed reporting variation; neither is inferred from a no-regret property.

\begin{corollary}[Resource margins and temperature]
\label{cor:margin_certificate}
Let $n\ge2$. Suppose that almost surely, for every user and round, a best candidate $k$ satisfies
\[
 r_{jk,t}-r_{j\iota_j,t}\ge g>0,\qquad
 r_{jk,t}-r_{ji,t}\ge\gamma>0\quad(i\ne k).
\]
The constants $g,\gamma$ are uniform, but $k$ may depend on the candidate realization. Then
\begin{equation}
 \mathsf{Gain}_T(n)\ge
 g-\frac{(n-1)e^{-\gamma/\nu}}{1+(n-1)e^{-\gamma/\nu}}.
 \label{eq:margin_gain_certificate}
\end{equation}
If Assumption~\ref{ass:amplitude} also holds, the following conditions are sufficient for $U_T(n)\ge B_T$:
\begin{equation}
 \frac{3\delta}{2g}\le\nu\le
 \frac{\gamma}{\log((n-1)(3-g)/g)},
 \qquad T\ge\frac{3n\Psi(\nu,\eta)}g.
 \label{eq:margin_temperature_budget}
\end{equation}
\end{corollary}

\begin{proof}
Fix a realized candidate vector and abbreviate its best value by $M$.
With $x=(n-1)e^{-\gamma/\nu}$, calibrated choice selects the best candidate with probability at least $1/(1+x)$. Since all values lie in $[0,1]$, its expected shortfall from $M$ is at most $x/(1+x)$. Subtract the original value and then average to obtain~\eqref{eq:margin_gain_certificate}.

For~\eqref{eq:margin_temperature_budget}, note that $0<g\le1$ and the logarithm is positive. Its temperature upper bound implies $x\le g/(3-g)$ and hence $x/(1+x)\le g/3$. The lower bound gives $\delta/(2\nu)\le g/3$, while the horizon condition gives $n\Psi(\nu,\eta)/T\le g/3$. Substitution in Theorem~\ref{thm:amplitude} proves the result.
\end{proof}

The temperature interval in~\eqref{eq:margin_temperature_budget} can be empty: its feasibility requires
\[
 \frac{3\delta}{2g}\log\frac{(n-1)(3-g)}g\le\gamma.
\]
This is a sufficient compatibility test between resource separation, market size, and reporting variation, not a necessary condition or an optimal temperature rule. The heterogeneous interval certificate can be substantially tighter because it retains each competitor's separation. Uniform positive margins exclude realizations in which no candidate improves on the original platform; they are not implied by the resource-center model.

\section{Verification Cost of Market Expansion}
\label{app:verification_proofs}

This appendix proves the cold-start obstruction in Section~\ref{sec:verification_cost}, extends it to other horizons, and separates it from the loss due to stochastic choice. Values and reports are constant throughout; the construction does not require changing descriptions or an alternating resource process.

\subsection{Cold-start bound for a general horizon}

We prove Proposition~\ref{thm:cold_start} through the following more general bound.

\begin{lemma}[Finite-horizon cold-start bound]
\label{lem:cold_start_horizon}
In the market of Proposition~\ref{thm:cold_start}, for every $\nu>0$ and $1\le T\le n-2$,
\begin{equation}
 U_T(n)\le\frac{1+c}{T}
       \bigl(H_{n-1}-H_{n-T-1}\bigr),
 \qquad H_k:=\sum_{s=1}^{k}\frac1s,\quad H_0:=0.
 \label{eq:cold_start_harmonic}
\end{equation}
\end{lemma}

\begin{proof}
All reports equal one, which is feasible under \eqref{eq:claims} by taking
$\lambda_i=1$; the platform with true value one has zero gap regardless of
its intensity. With $\eta=1$, a zero-value platform has score one until first
consumption and score zero thereafter. The value-one platform always has score
one. The original platform's score is one before its first consumption and $c$
thereafter.

Before round $t$, at most $t-1$ distinct zero-value platforms have been
consumed. For $1\le t\le n-1$, there are consequently at least
$n-t-1$ unverified zero-value platforms. Together with the value-one platform,
they contribute at least $(n-t)e^{1/\nu}$ to the softmax normalizer. Only the
value-one platform and the original platform contribute to its value-weighted
numerator, and their scores are at most one. Let $\mathcal F_t$ denote the
complete pre-choice analysis state; this does not require the UA to observe
unconsumed values. Conditional on that state,
\begin{equation}
 \mathbb E[r_{I_t,t}\mid\mathcal F_t]
 =\frac{\sum_i r_i e^{S_{i,t}/\nu}}{\sum_i e^{S_{i,t}/\nu}}
 \le\frac{(1+c)e^{1/\nu}}{(n-t)e^{1/\nu}}
 =\frac{1+c}{n-t}.
 \label{eq:cold_start_conditional_proof}
\end{equation}
No independence between successive choices or uniform coverage is used.
For $1\le T\le n-2$, take expectations and average to obtain
\begin{align}
 U_T(n)
 &\le\frac{1+c}{T}\sum_{t=1}^T\frac1{n-t}\notag\\
 &=\frac{1+c}{T}\bigl(H_{n-1}-H_{n-T-1}\bigr)
 \le\frac{1+c}{n-T}.
 \label{eq:cold_start_average_proof}
\end{align}
For $n\ge4$, the choice $T=\lfloor n/2\rfloor$ lies in this range and
$n-T\ge n/2$, giving \eqref{eq:cold_start_rate}. Since the old platform's
value is always $c$, $B_T=c$; the strict inequality follows when
$n>2(1+c)/c$. The best available value is one at every round, proving the
claimed constant resource-frontier gain.
The proof uses continued softmax competition between unverified reports and
the best platform; a different user-agent rule need not obey the bound.
\end{proof}

\subsection{Separating verification from stochastic choice}

At a fixed temperature, calibrated utility also decreases as the market expands in this construction. The following joint market--temperature family keeps calibrated choice nearly efficient and uses the same temperature for actual choice.

\begin{corollary}[Cold-start loss with nearly efficient calibrated choice]
\label{cor:cold_start_temperature}
Fix $c\in(0,1)$. For every integer $n\ge4$, use the market and policies of
Proposition~\ref{thm:cold_start}, with
\begin{equation}
 T_n=\lfloor n/2\rfloor,\qquad \nu_n=\frac{1}{2\log n}.
 \label{eq:cold_start_temperature_family}
\end{equation}
Evaluate actual and calibrated utility at this same temperature. Then
\begin{equation}
 1-U_{T_n}^0(n)\le(1-c)n^{-2(1-c)}+\frac1n,
 \qquad U_{T_n}(n)\le\frac{2(1+c)}n.
 \label{eq:cold_start_joint_bounds}
\end{equation}
Consequently, as $n\to\infty$,
\begin{equation}
 \mathsf{Distortion}_{T_n}(n)\to1,\qquad
 \mathsf{Gain}_{T_n}(n)\to1-c,\qquad
 U_{T_n}(n)-B_{T_n}\to-c.
 \label{eq:cold_start_joint_limits}
\end{equation}
\end{corollary}

The calibrated choice becomes nearly efficient while actual utility vanishes
over the same horizon and resource process. This is a joint parameter-family
counterexample, not a fixed-temperature large-market conclusion or a claim
that a fixed market fails forever. It also does not establish a dynamic
equilibrium. The limitation is specific to the stated mechanism: a different
user agent need not keep undiscovered reports competing with a discovered
value-one resource.

\begin{proof}
For the market in Proposition~\ref{thm:cold_start}, fully calibrated softmax
utility is constant across rounds and equals
\begin{equation}
 U_T^0(n)
 =\frac{e^{1/\nu}+c e^{c/\nu}}
        {e^{1/\nu}+e^{c/\nu}+n-2}.
 \label{eq:cold_start_calibrated_proof}
\end{equation}
Its shortfall from one satisfies
\begin{equation}
 1-U_T^0(n)
 =\frac{(1-c)e^{c/\nu}+n-2}
        {e^{1/\nu}+e^{c/\nu}+n-2}
 \le(1-c)e^{-(1-c)/\nu}+(n-2)e^{-1/\nu}.
 \label{eq:cold_start_calibrated_limit}
\end{equation}
Set $\nu=\nu_n=1/(2\log n)$. Then
$e^{-(1-c)/\nu_n}=n^{-2(1-c)}$ and $e^{-1/\nu_n}=n^{-2}$, so
\begin{equation}
 1-U_{T_n}^0(n)
 \le (1-c)n^{-2(1-c)}+\frac{n-2}{n^2}
 \le (1-c)n^{-2(1-c)}+\frac1n.
 \label{eq:cold_start_family_calibration_proof}
\end{equation}
Proposition~\ref{thm:cold_start} applies at every positive temperature and
therefore gives $U_{T_n}(n)\le2(1+c)/n$ at this same $\nu_n$.
Since $c\in(0,1)$ is fixed independently of $n$, the calibrated utility
converges to one and the nonnegative actual utility converges to zero.
Finally, $B_{T_n}=c$ for every $n$. Substitution into the definitions of
$\mathsf{Distortion}$ and $\mathsf{Gain}$ proves
\eqref{eq:cold_start_joint_limits}.

This comparison uses one joint sequence $(n,T_n,\nu_n)$ for both systems.
In contrast, for any fixed $\nu>0$, the numerator of
\eqref{eq:cold_start_calibrated_proof} is constant while its denominator
grows linearly in $n$, and the calibrated utility tends to zero. Thus neither
a fixed-temperature large-market conclusion nor interchange of long-run and
low-temperature limits is used.
\end{proof}

\subsection{Matching the order of the interaction budget}

\begin{corollary}[A linear budget in the same cold-start family]
\label{cor:cold_start_matching_budget}
Fix $c\in(0,1)$ and use the market and reporting profile of
Proposition~\ref{thm:cold_start}, with $\eta=1$ and
$\nu_n=1/(2\log n)$ for every $n\ge4$. For
$n>2(1+c)/c$ and $1\le T\le\lfloor n/2\rfloor$,
\begin{equation}
 U_T(n)\le\frac{2(1+c)}n<c=B_T.
 \label{eq:cold_start_short_budget}
\end{equation}
Conversely, for every fixed $C>1/(1-c)$, all sufficiently large $n$ satisfy
\begin{equation}
 T\ge Cn\quad\Longrightarrow\quad U_T(n)>B_T.
 \label{eq:cold_start_long_budget}
\end{equation}
Both statements concern expected cumulative-average utility under the same
mechanism, resources, and prescribed reporting profile.
\end{corollary}

\begin{proof}
For $T\le\lfloor n/2\rfloor$, Lemma~\ref{lem:cold_start_horizon} and
\eqref{eq:cold_start_average_proof} give
$U_T(n)\le(1+c)/(n-T)\le2(1+c)/n$. The stated restriction on $n$ makes this smaller than $c$.

In this market each gap is constant, so $\delta=0$.
Since $\eta=1$,
$\Psi(\nu_n,1)=1-e^{-1/\nu_n}=1-n^{-2}$.
Calibrated utility is independent of $T$, and
\eqref{eq:cold_start_calibrated_limit} gives, for every $T$,
\[
 \mathsf{Gain}_T(n)
 \ge1-c-(1-c)n^{-2(1-c)}-\frac1n.
\]
For $T\ge Cn$, Theorem~\ref{thm:amplitude} therefore implies
\begin{equation}
 U_T(n)-B_T
 \ge1-c-(1-c)n^{-2(1-c)}-\frac1n
       -\frac{1-n^{-2}}C.
 \label{eq:cold_start_matching_proof}
\end{equation}
Its right-hand side converges to $1-c-1/C>0$, which proves
\eqref{eq:cold_start_long_budget}. The convergence is uniform over horizons
$T\ge Cn$ because no remaining term depends on $T$.
\end{proof}

Thus the order $n$ of a sufficient interaction budget cannot be reduced
uniformly within this family: budgets $o(n)$ eventually belong to the
short-budget regime, whereas a large enough constant multiple of $n$
suffices. This does not identify the earliest crossing time, give a
per-user realized safety guarantee, or prove an information-theoretic
lower bound for other recommendation mechanisms. It matches only the
market-width order in the displayed joint temperature family; it does not
establish necessity of the separate $\eta$, $\nu$, or $\delta$ dependencies.

\section{Additional Model Details}
\label{app:dynamic_proofs}
\label{app:platform_policy}

The query and candidate process has a fixed joint law independent of description policies; independence across platforms, users, or rounds is not required. Before normalization or retrieval, raw queries and resources take the form $u_j+\xi$ and $c_i+\zeta$, with fixed perturbation laws satisfying $\mathbb E\xi=\mathbb E\zeta=0$. Thus the centers need not equal the means of normalized queries or retrieved candidates.

\begin{definition}[Admissible platform policy]
\label{def:platform_policy}
A policy $\pi_i$ chooses a possibly randomized $\lambda_{i,t}$ using the queries received, its own current candidates, and its past submissions and payoff feedback. It cannot condition on current UA choice shocks or future observations. Users' credibility records and competing platforms' resources are not disclosed to the platform.
\end{definition}

Non-disclosure does not rule out inference from available feedback. A unilateral deviation in $J_i^T$ regenerates subsequent choices, feedback, and credibility records rather than holding the realized history fixed.

\paragraph{Comparison with screening mechanisms.}
\label{app:mechanism_comparison}
For one user with fixed positive resource values, value-weighted engagement and selection count induce identical platform preferences. In this overlapping submodel, the mechanism of \citet[Appendix~A]{buening2024strategic} can detect misreports from consumed rewards without knowing the reporting map. Its incentive guarantee combines exclusion with greedy selection and a terminal uniform-allocation phase; these operations change the choice rule studied here. When values vary across users or rounds, the payoff equivalence no longer holds in general.

\section{Experimental Details}
\label{app:experiment_details}

\subsection{Shared interaction and evaluation protocol}
\label{app:experiment_protocol}

We use the mechanism and default settings of Sections~\ref{sec:model} and~\ref{sec:experiments}, with seeds 1200--1211. The without-assessment condition keeps $\omega\equiv0$. Nested markets reuse the first $n$ platforms of a maximum-width resource stream, retaining original assignments. Queries, local retrieval, PA sampling and UA choice use separate random streams. Across mechanisms, resources and sampling uniforms are paired, but learning actions adapt to their own feedback histories.

Saved horizons $T\in\{1,16,32,64,128,256,512,1024,2048,4096\}$ are cumulative prefixes, not independent replications or retuned learners. Parameter grids and representations are fixed before simulation.

\paragraph{Platform behavior.}
Fixed heterogeneous PAs draw $\lambda_i\sim\operatorname{Unif}[0.25,0.95]$ once. A maximal-reporting condition sets all intensities to one. Learning PAs use Exp3 over $\{0,0.25,0.5,0.75,1\}$. With weights $w_{ia,1}=1$ and full horizon $H=4096$, the probabilities and update are
\begin{equation}
 \begin{aligned}
 \gamma&=\min\left\{1,\sqrt{\frac{5\log5}{(\mathrm e-1)H}}\right\},\\
 \rho_{ia,t}&=(1-\gamma)\frac{w_{ia,t}}{\sum_b w_{ib,t}}+\frac\gamma5,\\
 \log w_{ia,t+1}&=\log w_{ia,t}
 +\frac\gamma5\frac{\mathbf1\{a=\ell_{i,t}\}(u_{i,t}/m)}{\rho_{ia,t}}.
 \end{aligned}
 \label{eq:experiment_exp3}
\end{equation}
Here $\ell_{i,t}$ is the sampled action and $u_{i,t}$ is the realized payoff in~\eqref{eq:platform_payoff}. PAs observe only their own aggregate payoffs, not competing resources, credibility records or counterfactual rewards. The finite grid is an experimental choice; these runs do not establish policy regret or convergence.

\paragraph{Estimation and uncertainty.}
Within each seed, utility averages $\sum_i p_{ji,t}r_{ji,t}$ over users and prefix rounds, conditional on sampled histories. This is an evaluator quantity: sampled choices generate PA rewards and credibility updates. All benchmarks use the same candidates; neither unconsumed values nor the reporting formula are supplied to the UA.

Net gains and mechanism differences are computed within seed. Pointwise $95\%$ intervals use $t_{11,0.975}$ times the standard error over 12 seeds, without multiplicity adjustment. Amazon intervals condition on its frozen data and representation; synthetic seeds also redraw centers and catalogs. Users, rounds and prefixes are not independent replications. Negative-net-gain user fractions describe within-seed heterogeneity, not failure probabilities for the population-average guarantee.

Fixed-policy contrasts isolate score correction; Exp3 contrasts include behavioral responses. Width sweeps also change resources and the payoff scale per platform, so they do not isolate verification speed. Full configurations and seed-level outputs are retained in the experiment artifacts.

\paragraph{Relation to the welfare guarantee.}
\label{app:experiment_theory_link}
The broad-market queries have full spherical support, local sampling permits repeated items, and Exp3 assigns positive probability to intensities zero and one. For broad-market Exp3 and $T\ge2$, no deterministic $\delta<1$ is therefore uniformly valid; fixed intensity instead has essential gap range $\lambda_i$. These runs test empirical behavior beyond the useful small-amplitude guarantee, not its certified budget. Observed ranges are not almost-sure certificates. Appendix~\ref{app:controlled_certificate_experiment} supplies a bounded-support test. Finite catalogs also need not satisfy the common-upper-tail assumptions, so measured frontier gains do not test their asymptotic rate.

\subsection{Synthetic resource generator}
\label{app:synthetic_details}

Write $\operatorname{unit}(v)=v/\|v\|_2$. User and platform centers are independent unit-normalized standard Gaussian vectors in $\mathbb R^8$. Each platform has 24 private items:
\begin{equation}
 v_{ik}=\operatorname{unit}(c_i+0.35z_{ik}),\qquad
 q_{j,t}=\operatorname{unit}(u_j+0.20\xi_{j,t}),\qquad
 z_{ik},\xi_{j,t}\sim\mathcal N(0,I_8).
 \label{eq:synthetic_generator}
\end{equation}
Items are drawn once; query innovations are fresh each round. For each query, each platform draws four local indices uniformly with replacement and submits the item with largest query cosine. Value is $(1+q_{j,t}^{\mathsf T}v_{a_{ji,t}})/2$. This fixed rule is independent of descriptions and sees no competing catalog. Centers are pre-normalization locations, not means of query-conditioned candidates.

The original platform is assigned before queries by
\begin{equation}
 \iota_j\in\arg\max_{i\in\{1,2\}}
 u_j^{\mathsf T}\left(\frac1{24}\sum_{k=1}^{24}v_{ik}\right).
 \label{eq:synthetic_original_platform}
\end{equation}
It supplies exactly the same subsequent candidates in the original and expanded markets and is not optimized retrospectively over all 32 platforms.

\subsection{Amazon data and semantic resources}
\label{app:amazon_data}

\paragraph{Data and representation.}
The Amazon Reviews 2023 Musical Instruments 5-core release contains 511,836 ratings, split globally by timestamp into approximately $80/10/10$ training, validation and test portions. We retain 2,048 text-covered training-eligible products and 256 users with at least three positive text-covered training interactions. The inherited product subset used MF-derived strata, but neither its old catalog labels nor MF predictions define the semantic market.

Frozen Qwen3-Embedding-0.6B~\citep{zhang2025qwenembedding} embeds titles, categories, features and descriptions, excluding reviews and rating statistics. Inputs are capped at 1,200 characters and 256 tokens; last non-padding token pooling and L2 normalization yield 1,024 dimensions. The cache covers 3,616 products, including profile items outside the candidate universe; its manifest records the model revision.

\paragraph{Platforms and original assignments.}
Euclidean KMeans forms 32 virtual catalogs. Among five predetermined starts with 40 Lloyd iterations and at least eight items per catalog, minimum inertia selects the partition before simulation. Catalog sizes range from 16 to 185; platform centers are empirical vector means, not observed seller attributes.

Each seed samples 128 users and permutes catalogs. User centers average up to eight most recent training products rated at least four stars, then normalize. As in~\eqref{eq:synthetic_original_platform}, the original platform maximizes profile--catalog affinity among the first two catalogs, using each catalog's size. This simulated assignment precedes queries and is retained at every width.

\paragraph{Queries and local retrieval.}
For the unit profile $u_j$ and $d=1024$, draw
\begin{equation}
 q_{j,t}=\operatorname{unit}\left(u_j+\frac{0.5}{\sqrt d}\xi_{j,t}\right),
 \qquad \xi_{j,t}\sim\mathcal N(0,I_d).
 \label{eq:amazon_query}
\end{equation}
Each platform samples four products uniformly with replacement from its own catalog, excluding products the user rated during training, and submits the best current-query cosine match. The simulated value is
\begin{equation}
 r_{ji,t}=\frac{1+\cos(q_{j,t},v_{a_{ji,t}})}2.
 \label{eq:amazon_text_value}
\end{equation}
\paragraph{Interpretation.}
The LLM supplies representations, not strategic text or feedback; consumption reveals the selected semantic proxy exactly. Generated outcomes are not observed consumer counterfactuals. Inherited sampling, full-period 5-core filtering, uncertain metadata timing and possible pretraining overlap limit interpretation. Held-out ratings do not tune the mechanism.

\subsection{Platform-policy comparisons}
\label{app:market_horizons}

\begin{table}[!htbp]
 \centering
 \begin{minipage}[t]{0.48\linewidth}
  \vspace{0pt}
  \centering
  \caption{Credibility assessment effects in synthetic markets.}
  \label{tab:synthetic_reputation}
  \small
  \setlength{\tabcolsep}{3pt}
  \begin{tabular}{@{}lrrr@{}}
  \toprule
  PA & \shortstack{Without\\assessment} & \shortstack{With\\assessment} & Change \\
  \midrule
  Fixed & $-0.007$ & $0.055$ & $0.062$ \\
  Maximal & $-0.062$ & $0.030$ & $0.092$ \\
  Exp3 & $-0.039$ & $0.048$ & $0.087$ \\
  \bottomrule
  \end{tabular}
 \end{minipage}\hfill
 \begin{minipage}[t]{0.48\linewidth}
  \vspace{0pt}
  \centering
  \caption{Credibility assessment effects in Amazon markets.}
  \label{tab:amazon_reputation}
  \small
  \setlength{\tabcolsep}{3pt}
  \begin{tabular}{@{}lrrr@{}}
  \toprule
  PA & \shortstack{Without\\assessment} & \shortstack{With\\assessment} & Change \\
  \midrule
  Fixed & $-0.013$ & $-0.008$ & $0.005$ \\
  Maximal & $-0.016$ & $-0.009$ & $0.007$ \\
  Exp3 & $-0.014$ & $-0.008$ & $0.007$ \\
  \bottomrule
  \end{tabular}
 \end{minipage}
 \par\smallskip
 \begin{minipage}{\linewidth}
  \footnotesize
  Seed means at $n=16$, $T=4096$, $\nu=0.1$ and $\eta=0.2$. Without assessment and With assessment give net utility relative to the original platform; Change is their paired difference before rounding. All paired pointwise 95\% intervals for Change exclude zero. Fixed denotes heterogeneous fixed intensities; Maximal sets all intensities to one. These two conditions hold PA actions unchanged across mechanisms, whereas Exp3 comparisons include platform responses. Amazon uses the semantic value proxy.
 \end{minipage}
\end{table}

Synthetic net-gain intervals are $[0.0455,0.0643]$ under fixed reporting and $[0.0390,0.0576]$ under Exp3; Amazon intervals are $[-0.0166,-0.0001]$ and $[-0.0160,0.0004]$. Thus positive paired assessment benefits need not imply benefit over the original platform. Negative-net-gain user fractions average $29.2\%/33.1\%$ for synthetic fixed/Exp3 and $67.5\%/66.5\%$ for Amazon.

\subsection{Choice-temperature sensitivity}
\label{app:amazon_sensitivity}

\begin{table}[t]
 \centering
 \small
 \setlength{\tabcolsep}{5pt}
 \caption{Choice-temperature sensitivity with Exp3 PAs and credibility assessment, at $n=16$, $T=4096$ and $\eta=0.2$. Value columns are in units of $10^{-3}$. Means and pointwise 95\% intervals for actual net gain use 12 seeds.}
 \label{tab:temperature_sensitivity}
 \begin{tabular}{@{}lrrrrr@{}}
 \toprule
 Market & $\nu$ & \shortstack{Calibrated\\gain}
 & \shortstack{Actual\\net gain} & 95\% CI & Distortion \\
 \midrule
 Synthetic & $0.03$ & $165.2$ & $99.8$ & $[89.8,109.8]$ & $65.4$ \\
           & $0.10$ & $92.5$ & $48.3$ & $[39.0,57.6]$ & $44.2$ \\
           & $0.30$ & $4.4$ & $-11.9$ & $[-20.3,-3.6]$ & $16.3$ \\
 \midrule
 Amazon    & $0.03$ & $12.8$ & $6.1$ & $[-1.9,14.1]$ & $6.7$ \\
           & $0.10$ & $-6.6$ & $-7.8$ & $[-16.0,0.4]$ & $1.2$ \\
           & $0.30$ & $-12.9$ & $-13.3$ & $[-21.8,-4.9]$ & $0.4$ \\
 \bottomrule
 \end{tabular}
\end{table}

Across the tested temperatures, lower values improve calibrated and actual mean gains in both markets under Exp3, despite larger distortion (Table~\ref{tab:temperature_sensitivity}). Better selection can outweigh increased reporting sensitivity, but Amazon's actual net-gain interval at $\nu=0.03$ still crosses zero. Resources are reused across temperatures; PA actions adapt to each UA's feedback history.

\subsection{Controlled certificate experiments}
\label{app:controlled_certificate_experiment}

\paragraph{Shared bounded-support construction.}
Each platform submits $a_i=\boldsymbol e_i$. User $j$'s query center has coordinates $0.4$ at its original platform $1$, $h$ at a preferred platform $i_j^\star\ne1,2$, and $0.1$ elsewhere. Queries add independent coordinate noise uniform on $[-0.01,0.01]$, with $r(q,a)=\operatorname{clip}_{[0,1]}(q^{\mathsf T}a)$. Candidates and query laws are fixed. Each PA selects one intensity in $[1-a,1]$ per round, shared across users; this restriction is imposed, not induced by competition. Both protocols use 128 users, 4,096 rounds, 12 seeds, $\nu=0.1$ and $\eta=0.2$.

\paragraph{Support-based certificate.}
The value supports are $[0.39,0.41]$, $[h-0.01,h+0.01]$ and $[0.09,0.11]$ for the original, preferred and remaining platforms. Proposition~\ref{prop:support_certificate} gives
\begin{align}
 \delta&=0.02+0.89a,\notag\\
 G&=0.09-0.41+
 \frac{h-0.1}{1+\exp((0.42-h)/0.1)+(n-2)\exp((0.12-h)/0.1)}.
 \label{eq:experimental_support_certificate}
\end{align}
The widest gap interval is $[(1-a)0.89,0.91]$, at a low-value platform. The reciprocal denominator bounds the preferred platform's calibrated selection probability using its lowest value and competitors' highest values. Every resource is worth at least $0.09$, the preferred resource adds at least $h-0.1$, and the original is worth at most $0.41$. Thus $G$ is an analytical lower bound, not a measured gain.

For $G>\delta/(2\nu)$, Corollary~\ref{cor:no_drop} gives the certified budget
\[
 T_{\mathrm{cert}}=\left\lceil
 \frac{n\Psi(\nu,\eta)}{G-\delta/(2\nu)}
 \right\rceil.
\]
At $(n,h,a)=(16,0.9,0.05)$, $\delta=0.0645$, $G>0.468981$ and $\Psi(0.1,0.2)<13.405602$ by~\eqref{eq:nonlinear_budget_tail}. Substitution gives a certified budget of 1,465 rounds and net gain above $0.0941$ at $T=4096$. All certificates use analytical supports and conservative rounding, not observed extrema.

\paragraph{Restricted Exp3.}
The first protocol fixes $(n,h,a)=(16,0.9,0.05)$, with preferred identities cyclically assigned to platforms $3,\ldots,16$, and runs~\eqref{eq:experiment_exp3} on $\{0.95,\allowbreak 0.9625,\allowbreak 0.975,\allowbreak 0.9875,\allowbreak 1\}$. Final net gain is $0.4658$, with interval $[0.4657,0.4659]$, versus $-0.2210$ without assessment. The first saved positive interval occurs at 128 rounds, while the 64-round interval is negative. This observed transition is not a stopping-time guarantee; the analytical budget covers every admissible policy in the interval.

\paragraph{Market, resource and reporting sweeps.}
\label{app:certificate_slices}
The second protocol retains preferred platforms 3 or 4 at every width. Intensities are drawn independently from $\operatorname{Unif}[1-a,1]$ each round, independently of UA actions: these are open-loop, not learning policies. Table~\ref{tab:certificate_slices} varies one parameter around $(16,0.9,0.05)$, giving seven settings and 84 seeded runs.

\begin{table}[t]
 \centering
 \small
 \setlength{\tabcolsep}{8pt}
 \renewcommand{\arraystretch}{1.15}
 \caption{Observed net utility and theoretical guarantees in the controlled synthetic market at $T=4096$. The base is $(n,h,a)=(16,0.9,0.05)$; other rows change one parameter. Lower bounds are rounded down. All observed-net-gain interval half-widths are below $0.0001$ over 12 seeds. A dash means no finite positive-gain budget is certified.}
 \label{tab:certificate_slices}
 \begin{tabular}{@{}lrrrr@{}}
 \toprule
 Setting & \shortstack{Calibrated\\gain} & \shortstack{Observed\\net gain} & \shortstack{Certified\\lower bound} & \shortstack{Certified\\rounds} \\
 \midrule
 Base & $0.493$ & $0.466$ & $0.094$ & 1,465 \\
 \midrule
 $n=4$ & $0.496$ & $0.491$ & $0.137$ & 357 \\
 $n=32$ & $0.489$ & $0.433$ & $0.036$ & 3,034 \\
 \addlinespace
 $h=0.60$ & $0.140$ & $0.120$ & $-0.305$ & --- \\
 $h=0.75$ & $0.327$ & $0.301$ & $-0.084$ & --- \\
 \addlinespace
 $a=0$ & $0.493$ & $0.465$ & $0.316$ & 582 \\
 $a=0.10$ & $0.493$ & $0.466$ & $-0.129$ & --- \\
 \bottomrule
 \end{tabular}
\end{table}

Entrants do not improve the best resource, so the width sweep combines choice dilution and verification effects. Increasing $a$ also lowers mean intensity, not just widening its range. Observed gains remain positive even when no finite budget is certified.

\subsection{Alternative calibration with matched forgetting}
\label{app:matched_calibration}

We compare~\eqref{eq:reputation} with an affine gap predictor using matched forgetting, initialization and score range. For one user--platform record, index verified consumptions by $s=1,\ldots,k$, with reports $A_s$, gaps $d_s=A_s-r_s$ and $\rho=1-\eta$. Before its next consumption, fit
\begin{equation}
 (\hat a_k,\hat b_k)\in\arg\min_{a,b}
 \left\{\rho^k a^2+
 \eta\sum_{s=1}^k\rho^{k-s}(d_s-a-bA_s)^2+\kappa b^2\right\},
 \qquad \kappa=0.01.
 \label{eq:matched_affine_gap}
\end{equation}
The score is $A-\operatorname{clip}_{[0,1]}(\hat a_k+\hat b_kA)$, with zero initial coefficients and the same temperature as EMA. Only the consumed record updates, at round end; unconsumed records do not decay. Neither intensities nor the reporting formula are supplied, and regularization is fixed in advance. Restricting $b=0$ recovers the zero-initialized EMA exactly. The extension uses the current report as a predictor, unlike lagged EMA.

Comparisons reuse the original resources and seeds. Fixed-policy comparisons share actions; Exp3 populations learn separately under each UA. Tables~\ref{tab:matched_calibration} and~\ref{tab:amazon_matched_calibration} report paired differences.

\begin{table}[!htbp]
 \centering
 \begin{minipage}[t]{0.48\linewidth}
  \vspace{0pt}
  \centering
  \caption{Calibration in synthetic markets.}
  \label{tab:matched_calibration}
  \small
  \setlength{\tabcolsep}{4pt}
  \begin{tabular}{@{}lrr@{}}
  \toprule
  PA & Mean & 95\% CI \\
  \midrule
  Fixed & $4.94$ & $[4.86,5.02]$ \\
  Exp3 & $-0.32$ & $[-0.47,-0.16]$ \\
  Restricted Exp3 & $22.82$ & $[22.74,22.90]$ \\
  \bottomrule
  \end{tabular}
 \end{minipage}\hfill
 \begin{minipage}[t]{0.48\linewidth}
  \vspace{0pt}
  \centering
  \caption{Calibration in Amazon markets.}
  \label{tab:amazon_matched_calibration}
  \small
  \setlength{\tabcolsep}{4pt}
  \begin{tabular}{@{}lrr@{}}
  \toprule
  PA & Mean & 95\% CI \\
  \midrule
  Fixed & $0.13$ & $[0.11,0.15]$ \\
  Exp3 & $-0.29$ & $[-0.32,-0.25]$ \\
  \bottomrule
  \end{tabular}
 \end{minipage}
 \par\smallskip
 \begin{minipage}{\linewidth}
  \footnotesize
  Entries report affine gap calibration minus EMA utility in $10^{-3}$ units; positive values favor affine calibration. Both use matched forgetting, with $m=128$, $n=16$ and $T=4096$. CIs are paired pointwise 95\% intervals over 12 seeds. Fixed denotes heterogeneous fixed intensities; Exp3 uses the full action grid. Restricted Exp3 uses the controlled synthetic market. Amazon values use the semantic outcome proxy.
 \end{minipage}
\end{table}

Neither update uniformly dominates. Affine calibration performs better under fixed reporting and restricted Exp3; broad-market Exp3 slightly favors EMA. These comparisons test sensitivity to predictor form, not optimality among all same-feedback methods.

\subsection{Observed ratings and outcome-model sensitivity}
\label{app:amazon_outcome_validation}

\paragraph{Observed-rating check.}
Using the same chronological split (timestamp ties enter the later split), an existing diagnostic covers 130 test ratings from 58 eligible users. Training-profile cosine, used because review-time queries are unobserved, has Spearman correlation $0.101$ with ratings, with interval $[-0.117,0.265]$. Restricting to unseen pairs in the 2,048-product universe leaves 19 ratings from 15 users.

A broader cohort requires at least ten training ratings overall and three positive training ratings within that universe; profiles use up to eight latest positives. Of 1,717 eligible users, 92 contribute 119 unseen test ratings. Correlation is $-0.185$, with interval $[-0.349,-0.006]$. Both 95\% intervals use 2,000 user-cluster bootstrap draws, conditional on frozen data and representation. These sparse diagnostics are not new untouched tests or validation of consumer satisfaction; missing ratings are not negative outcomes.

\paragraph{A separate rating-derived outcome proxy.}
We retain the Qwen catalogs, queries, retrieval, candidates and original assignments, but replace relevance by a frozen 32-dimensional biased matrix-factorization prediction, clipped to $[0,1]$. The outcome model is trained on training ratings and selected by validation RMSE; it is not a second recommendation-side embedding. Reports, feedback, PA payoff and welfare all use the replacement. The UA sees only consumed outcomes, not the predictor; the original platform is not reoptimized.

\paragraph{Results and scope.}
At $n=16$, $m=128$, $T=4096$, $\nu=0.1$ and $\eta=0.2$, assessment improves utility over uncorrected reports by $0.00330$ under fixed reporting and $0.00225$ under Exp3, with paired intervals $[0.00238,0.00421]$ and $[0.00197,0.00253]$. Net gains over the original platform are $0.01314$ and $0.00858$, with intervals $[0.00679,0.01949]$ and $[0.00238,0.01477]$. Correction benefits persist, but the positive net gains contrast with the semantic-proxy results: their sign depends on the outcome model.

The rating proxy is not demonstrably better: its normalized RMSE on the complete warm-start test split is $0.2951$, versus $0.2948$ for a bias-only model. Predictions are not observed counterfactuals, and inherited product selection makes the two proxies non-independent market samples. The data limitations in Appendix~\ref{app:amazon_data} remain. The mechanism intervals measure variability over 12 simulation seeds conditional on frozen data and models, not uncertainty about real consumer welfare.

\end{document}